\documentclass[10pt,twocolumn,letterpaper]{article}

\usepackage[pagenumbers]{cvpr} 

\usepackage{amsthm, amssymb}
\usepackage[linesnumbered,ruled,vlined]{algorithm2e}

\usepackage{multirow}
\usepackage{graphicx}
\usepackage{wrapfig}
\usepackage{lineno}

\allowdisplaybreaks
\theoremstyle{definition}
\newtheorem{definition}{Definition}[section]

\definecolor{cvprblue}{rgb}{0.21,0.49,0.74}
\usepackage[pagebackref,breaklinks,colorlinks,allcolors=cvprblue]{hyperref}

\def\paperID{715} 
\def\confName{CVPR}
\def\confYear{2025}

\title{Ferret: An Efficient Online Continual Learning Framework under Varying Memory Constraints}

\author{
Yuhao Zhou$^{1}$\quad
Yuxin Tian$^{1}$\quad 
Jindi Lv$^{1}$\quad
Mingjia Shi$^{2}$\quad 
Yuanxi Li$^{3}$\quad
\\
Qing Ye$^{1}$\footnotemark[2]\quad
Shuhao Zhang$^{4}$\quad
Jiancheng Lv$^{1}$\quad
\\
$^{1}$Sichuan University\;
\quad
$^{2}$National University of Singapore\;
\quad
\\
$^{3}$University of Illinois Urbana-Champaign\;
\quad
$^{4}$Huazhong University of Science and Technology\;
}

\begin{document}
\maketitle
\begin{abstract}
In the realm of high-frequency data streams, achieving real-time learning within varying memory constraints is paramount. 
This paper presents Ferret, a comprehensive framework designed to enhance \emph{online accuracy} of Online Continual Learning (OCL) algorithms while dynamically adapting to varying memory budgets.
Ferret employs a fine-grained pipeline parallelism strategy combined with an iterative gradient compensation algorithm, ensuring seamless handling of high-frequency data with minimal latency, and effectively counteracting the challenge of stale gradients in parallel training. 
To adapt to varying memory budgets, its automated model partitioning and pipeline planning optimizes performance regardless of memory limitations. 
Extensive experiments across 20 benchmarks and 5 integrated OCL algorithms show Ferret’s remarkable efficiency, achieving up to 3.7$\times$ lower memory overhead to reach the same online accuracy compared to competing methods.
Furthermore, Ferret consistently outperforms these methods across diverse memory budgets, underscoring its superior adaptability. 
These findings position Ferret as a premier solution for efficient and adaptive OCL framework in real-time environments.
\end{abstract}    
\section{Introduction}
\label{sec:intro}

Data is crucial for Machine Learning (ML), forming the basis for algorithms and models~\cite{touvron2023llama, girdhar2023imagebind, menghani2023efficient}.
In real-world applications~\cite{david2022tweets, dixon2020machine}, data arrives in high-frequency streams with varying distributions~\cite{mai2022online, wang2024comprehensive}.
This makes data time-sensitive and short-lived~\cite{cai2021online, lin2021clear, valavi2022time}, rendering offline-trained models based on historical data ineffective for future data of unknown distribution~\cite{niu2022efficient}.
Thus, the significance of \emph{Online Continual Learning} (OCL) is growing~\cite{jayarajan2021lifestream, barber2021bladerunner, veseli2023streaming}, as it enables learning over data streams to adapt to dynamic data distributions in real-time.

In the literature, OCL tackles two main challenges: 1) mitigating catastrophic forgetting~\cite{lopez2017gradient}, where the model retains previously learned knowledge while acquiring new information (\textit{e.g.}, regularization-based~\cite{buzzega2020dark, caccia2021new, de2021continual2, aljundi2018memory}, replay-based~\cite{chaudhry2019continual, isele2018selective, rebuffi2017icarl}, sampling-based~\cite{aljundi2019online, aljundi2019gradient, yoon2021online}, others~\cite{fernando2017pathnet, prabhu2020gdumb}, etc.), and 2) enhancing rapid adaptation~\cite{cai2021online, lin2021clear}, which involves swiftly adjusting to new data or tasks (\textit{e.g.}, latency-oriented~\cite{sahoo2017online, ghunaim2023real}, buffering~\cite{yoon2017lifelong, mirzasoleiman2020coresets}, others~\cite{rusu2016progressive, finn2019online}, etc.).
In general, the increasing demand for resource-limited systems that can seamlessly integrate new information with minimal latency has driven the popularity of OCL~\cite{gunasekara2023survey}
Therefore, this paper explores the challenge of rapid adaptation under varying memory constraints in OCL.

To effectively address the above OCL challenge, it is essential to explore solutions beyond mentioned algorithmic improvements by also optimizing the underlying framework.
An efficient OCL framework must prioritize both processing speed and memory management under the limited memory capacity so that it can efficiently handle unlimited data streams with dynamic data distributions for increased \emph{online accuracy}~\cite{cai2021online} (\textit{i.e.}, a metric measuring real-time accuracy for continuous new data predictions).
Specifically, the framework should quickly process incoming data to extract valuable insights and make informed decisions~\cite{tai2018sketching, muhammad2020deep} by minimizing both the latency from data receipt to its initial processing and the time taken for the learning process itself.
Additionally, the framework is not only expected to operate within a predetermined memory allotment but also to demonstrate scalability across diverse memory capacities~\cite{de2018distributed, murshed2021machine}.
This duality ensures that the framework remains efficient regardless of the memory resources available, thereby maintaining consistent performance in dynamic environments.

Numerous ML frameworks have been proposed that offer innovative approaches to scalable and flexible ML development~\cite{kim2016strads, moritz2018ray, darvish2019deepsigns, zheng2019hiwaylib, hu2020deepsniffer, zhang2020retiarii, hu2023hydro, guan2023amanda, ansel2024pytorch}.
For instance, Ray~\cite{moritz2018ray} facilitates distributed computing on any scale, while Pytorch~\cite{ansel2024pytorch} excels in dynamic computation graphs for model training.
Despite their advancements, these frameworks often do not specifically address the unique requirements of learning over streaming data~\cite{ghunaim2023real}, which is a key focus of OCL.
Recently, there are some frameworks dedicated to OCL by prioritizing real-time data processing~\cite{xie2020kraken, li2022camel, kwon2023lifelearner}.
Nevertheless, they either lack general applicability or fail to balance processing speed with consumed memory, leading to reduced online accuracy and low memory scalability, underscoring the need for innovative solutions in this domain.

In this work, we propose an OCL framework named Ferret, designed to achieve e\underline{\textbf{F}}fici\underline{\textbf{E}}nt pipeline lea\underline{\textbf{R}}ning over f\underline{\textbf{R}}equ\underline{\textbf{E}}nt data s\underline{\textbf{T}}reams for enhanced online accuracy across memory constraints.
Ferret comprises a fine-grained pipeline parallelism component with an iterative gradient compensation algorithm and a model partitioning and pipeline planning component.
Firstly, to facilitate rapid adaptation over frequent streaming data for higher online accuracy, Ferret employs a fine-grained pipeline parallel strategy, allowing precise control over each pipeline stage for seamless data management.
Additionally, to mitigate the impact of stale gradients in parallel processing, Ferret integrates a novel iterative gradient compensation algorithm.
Secondly, to guide the selection of optimal model partition schemes and pipeline configurations under given memory budgets, Ferret solves the involved multivariate optimization problem through a bi-level optimization algorithm.

Our contributions can be outlined as follows:

\begin{itemize}
    \item We propose a framework named Ferret for boosting the online accuracy of OCL algorithms under memory constraints.
    To the best of our knowledge, this is the first work focusing on enhancing OCL by employing pipeline parallelism and scheduling.
    \item To process high-frequent data streams without delay, Ferret employs a fine-grained pipeline parallelism strategy, enabling interleaved processing of incoming streaming data. 
    Furthermore, Ferret utilizes an iterative gradient compensation algorithm to efficiently mitigate the effects of stale gradients across different pipeline stages, preventing performance degradation.
    \item We derive the optimal parameters for automatic model partitioning and pipeline planning by mapping the involved multi-variable optimization problem into a bi-level optimization problem.
    \item Extensive experiments on 20 benchmarks demonstrate that our proposed framework consistently enables more efficient OCL within given memory budgets.
    The code is open-sourced for reproduction.
\end{itemize}

\section{Related Work}
\label{sec:related}

The current OCL research focuses on two areas: mitigating catastrophic forgetting and enhancing rapid adaptation.

\textbf{Mitigating catastrophic forgetting: }
Catastrophic forgetting, often quantified by the test accuracy~\cite{li2022camel, ghunaim2023real, lin2021clear}, poses a significant barrier to the efficacy of OCL in dynamic environments, where the ability to preserve historical information is crucial.
Multiple directions have emerged to reduce catastrophic forgetting, including: 
1) regularization-based techniques~\cite{buzzega2020dark, caccia2021new, de2021continual2, aljundi2018memory} impose constraints on weight updates to preserve important parameters that are crucial for past tasks.
2) replay-based techniques~\cite{chaudhry2019continual, isele2018selective, rebuffi2017icarl} help the model to rehearse old knowledge alongside new information by maintaining a memory of previous data.
3) sampling-based techniques~\cite{aljundi2019online, aljundi2019gradient, yoon2021online} enhance the efficiency of replay mechanisms by selectively choosing the most relevant data samples for rehearsal.
4) other techniques~\cite{fernando2017pathnet, prabhu2020gdumb} focus on various novel approaches, such as modular networks and dynamically allocated resources, to protect previously learned information from being overwritten.

\textbf{Enhancing rapid adaptation:}
Rapid adaptation is in scenarios where immediate processing of incoming data is required~\cite{cai2021online, lin2021clear}, which is often quantified by the online accuracy~\cite{cai2021online} defined as $oacc_\mathcal{A}(t) = \sum_{i=1}^t acc(y^i, \hat{y}^i) / t$.
Strategies developed to enhance rapid adaptation include:
1) latency-oriented techniques~\cite{sahoo2017online, ghunaim2023real} iteratively generate predictions and update model parameters immediately upon the arrival of streaming data by discarding data that cannot be processed in time.
2) buffering-oriented techniques~\cite{yoon2017lifelong, mirzasoleiman2020coresets} buffers and samples incoming data streams and apply periodic batch-training~\cite{cuda}.
3) other techniques~\cite{rusu2016progressive, finn2019online} introduce novel methods like adapting model structures in response to new tasks and learning how to learn efficiently, to adapt to dynamic data distributions rapidly.

\section{Motivation}

To effectively navigate the challenges posed by OCL, it is crucial to expand our approach beyond merely refining mentioned OCL algorithms, by also enhancing the underlying ML framework to adaptively balance processing speed with efficient memory management under memory constraints.
Particularly, boosting processing speed is essentially reducing data processing time, which can be represented as $t^l + F / R_h P_h$, where $t^l$ denotes the latency from data arrival to processing, $F$ denotes the required floating point operations (FLOPS) by the underlying OCL algorithm, $R_h$ and $P_h$ denote the hardware utilization rate and the theoretical floating point operations per second (FLOPs) of the hardware, respectively.
Clearly, only $t^l$ and $R_h$ are optimizable by the framework.

Existing ML frameworks mainly focus on: 1) distributed and parallel computing~\cite{kim2016strads, moritz2018ray, zheng2019hiwaylib}, 2) Optimized model training and deployment~\cite{zhang2020retiarii, hu2023hydro, ansel2024pytorch}, and 3) others including security~\cite{hu2020deepsniffer, darvish2019deepsigns} and debugging~\cite{guan2023amanda}.
These frameworks facilitate scalable and flexible ML, yet they rarely tackle the challenges of managing streaming data.
Regrettably, the few ML frameworks designed for OCL~\cite{xie2020kraken, li2022camel, kwon2023lifelearner} either lack general applicability or fail to concurrently optimize $t^l$ and $R_h$ within memory limitations.
For instance, Kraken~\cite{xie2020kraken} is tailored for recommendation systems.
Conversely, while Camel~\cite{li2022camel} and LifeLearner~\cite{kwon2023lifelearner} boost $R_h$ via buffering and sampling, they also raise $t^l$ and memory usage, reducing online accuracy.

Pipeline parallelism can naturally process sequential streaming data while utilizing batch training.
This motivates us to incorporate pipeline parallelism into OCL to simultaneously minimize $t^l$ and maximize $R_h$ under a given memory budget, thereby boosting online accuracy.
We achieve this balance through refined scheduling strategies and better hardware integration, ensuring optimal resource utilization within the constraints of memory budgets.

\section{Problem Formulation}
In this section, we define the problem we aim to address. Note that the notations used throughout this paper are defined in Sec.~\ref{sec:notation} in the appendix.

Consider a general learning problem defined over a feature space $\mathcal{X}$ and a label space $\mathcal{Y}$ that aims to minimize a loss function $\mathcal{L}(D^t; \theta)$ where data $D^t = (\mathbf{x}^t, \mathbf{y}^t) \in \mathcal{X} \times \mathcal{Y}$ arrives at timestamp $t$.
Our objective is to rapidly derive an updated model $\theta^{t}$ with $D^t$ and $\theta^{t-1}$ under a given memory constraint $M$, so that the online accuracy of $\theta^{t}$ is high.
Unlike updating a model offline with a pre-collected dataset, $D^t$ will be discarded after updating $\theta^{t-1}$ in OCL.

Directly optimizing online accuracy in our objective during runtime is hard, as the online accuracy is only calculable after obtaining labels of incoming data.
Instead, we measure the volume of data values learned by the model as a proxy to estimate and optimize online accuracy.
Formally, assuming $D^t$ has an initial data value of $V_{D^t}$ and its data value declines as a time-dependent exponential decay function~\cite{valavi2022time}, we define the Adaptation Rate as follows.
\begin{definition}[Adaptation Rate of A OCL framework]
Consider a OCL framework $\mathcal{A}$ receives a data $D^t = (\mathbf{x}^t, \mathbf{y}^t)$ at timestamp $t$ that has an initial data value of $V_{D^t}$, and updates a model $\theta^{t-1}$ in the hypothesis space $\Theta$ at timestamp $t + r_\mathcal{A}^t$ ($r_\mathcal{A}^t = +\infty$ if $D^t$ is discarded).
Let the data value of $D^t$ decline as a time-dependent exponential decay function, and new data $D^t$ constantly arrives until $t=T$.
The Adaptation Rate of $\mathcal{A}$ is defined as
{\small  
  \setlength{\abovedisplayskip}{6pt}
  \setlength{\belowdisplayskip}{\abovedisplayskip}
  \setlength{\abovedisplayshortskip}{0pt}
  \setlength{\belowdisplayshortskip}{3pt}
\begin{equation}
    R_{\mathcal{A}}^T = \frac{\sum_{t=0}^T e^{-c r_\mathcal{A}^t}V_{D^t}}{T},
\end{equation}
}
where the constant $c$ describes the reduction rate of $V_{D^t}$.
\label{def:vre}
\end{definition}

With Def.~\ref{def:vre}, our objective can be formulated as
{\small  
  \setlength{\abovedisplayskip}{6pt}
  \setlength{\belowdisplayskip}{\abovedisplayskip}
  \setlength{\abovedisplayshortskip}{0pt}
  \setlength{\belowdisplayshortskip}{3pt}
\begin{equation}
    \max_\mathcal{A} \mathcal{R}_\mathcal{A}^T~~\mathrm{s.t.}~~\mathcal{M}_\mathcal{A} \leq M,
\label{eq:main-obj}
\end{equation}
}
where $\mathcal{M}_\mathcal{A}$ is the memory footprint of $\mathcal{A}$ during training.
\section{Methodology}
\label{sec:methodology}

\begin{figure*}
  \begin{minipage}{0.79\textwidth}
    \centering
    \includegraphics[width=1.0\textwidth]{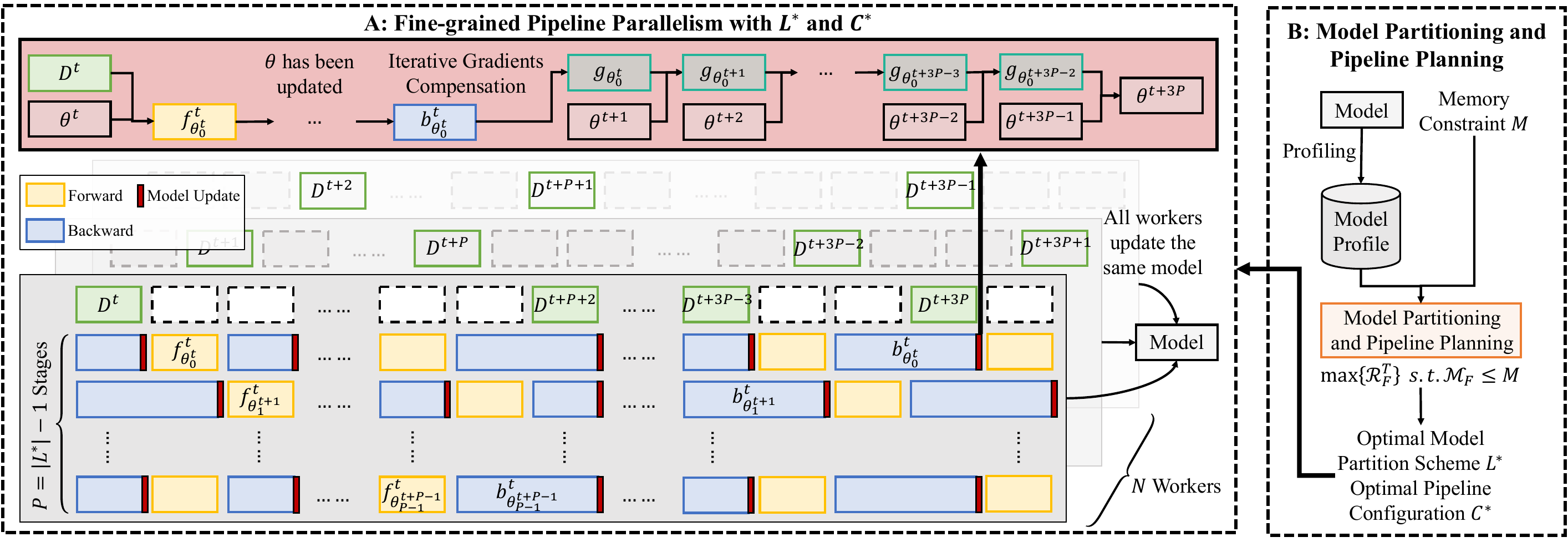}
    \captionof{figure}{
    The overall workflow of Ferret. 
    In A, based on the optimal model partition scheme $L^*$ and pipeline configuration $C^*$, $N$ workers are spawned to initiate fine-grained pipeline parallelism that consumes streaming data interleavedly, and update the same model asynchronously by iteratively compensating stale gradients.
    In B, $L^*$ and $C^*$ are obtained by optimizing Eq.~\ref{eq:main-obj}. 
    }
    \label{fig:overview}
  \end{minipage}%
  \hfill
  \begin{minipage}{0.2\textwidth}
    \centering
    \captionsetup[subfloat]{labelfont=scriptsize,textfont=scriptsize}
    \subfloat[Activation Recomputation (T1)\label{fig:prune-ar}]{\includegraphics[width=1.0\linewidth]{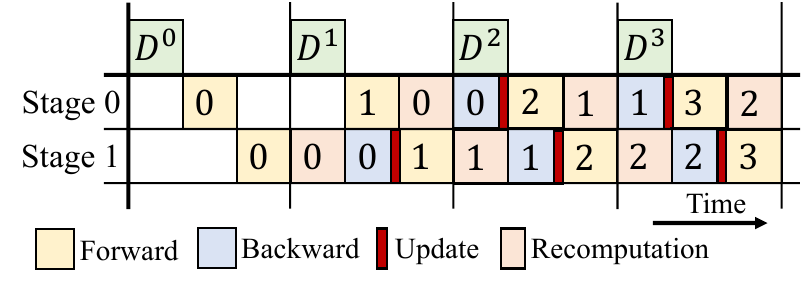}}
    \\
    \subfloat[Gradient Accumulation (T2)\label{fig:prune-ga}]{\includegraphics[width=1.0\linewidth]{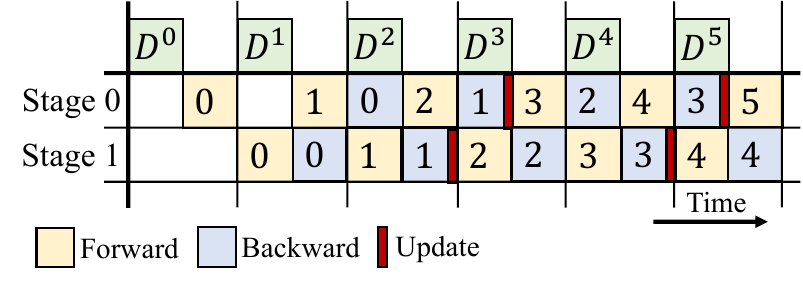}}
    \\
    \subfloat[Backpropagation Omission (T3)\label{fig:prune-bo}]{\includegraphics[width=1.0\linewidth]{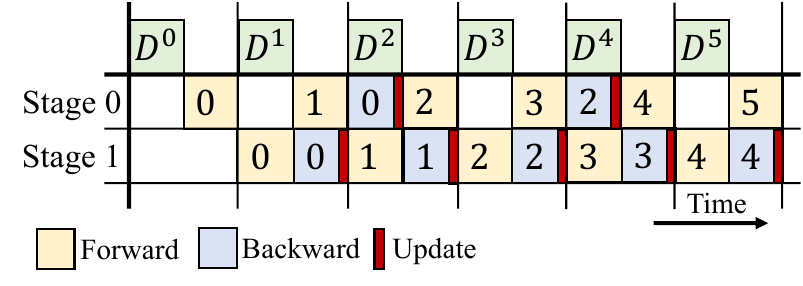}}
    \\
    \subfloat[Worker Removal (T4)\label{fig:prune-wr}]{\includegraphics[width=1.0\linewidth]{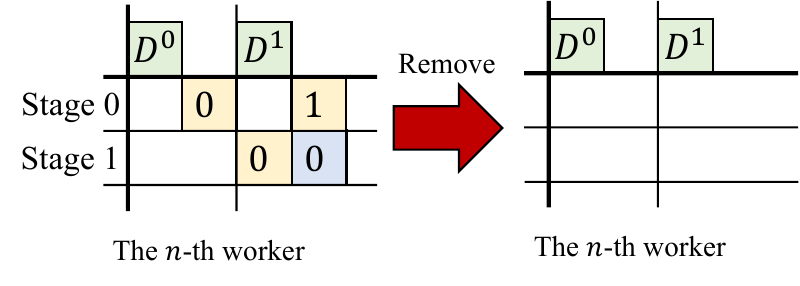}}
  \end{minipage}
\end{figure*}

The workflow of Ferret is shown in Fig.~\ref{fig:overview}, comprising a fine-grained pipeline parallelism component (A), followed by a model partitioning and pipeline planning component (B).
In A, the model is trained using Ferret's fine-grained pipeline parallelism to manage high-frequent data streams with minimal latency.
Given the high degree of parallelism within the system, gradient staleness can become significant and variable, potentially causing severe model degradation.
To mitigate this issue, an iterative gradient compensation algorithm is applied prior to model updating.
In B, the model is profiled to optimize Eq.~\ref{eq:main-obj},  determining the optimal model partition scheme and pipeline configuration.

\subsection{Fine-grained Pipeline Parallelism}

\subsubsection{Architectural design}

Ferret utilizes an asynchronous pipeline parallelism strategy with 1F1B scheduling to process streaming data immediately upon arrival.
To efficiently handle high-frequency data streams without delay, it is imperative that $t^f + t^b$ is minimized.
However, $t^f + t^b$ is inherently lower bounded by $(\max_i \hat{t}_i^f + \max_i \hat{t}_i^b)$, indicating that some of the data must be discarded if $t^d$ is less than this lower bound.
To prevent the loss of data, Ferret enhances system throughput by deploying $N \leq \lceil (t^f + t^b)/t^d \rceil$ workers, each performing pipeline parallelism concurrently over interleaved data streams. 
Specifically, the $i$-th data is processed by the $n$-th worker if and only if $i \equiv c_n^d \mod \lceil (t^f + t^b)/t^d \rceil$.
This strategy, while effective in reducing latency, significantly increases memory usage.
Therefore, Ferret balances the trade-offs between $\mathcal{R}$ and $\mathcal{M}$ by collectively employing four \textbf{t}echniques: activation recomputation~\cite{chen2016training}, gradient accumulation~\cite{narayanan2021memory}, back-propagation omission and worker removal, allowing precise control over each pipeline stage for seamless data management.

\textbf{T1. Activation Recomputation:} 
Activation recomputation exchanges additional computational overhead for reduced memory usage, as Fig.~\ref{fig:prune-ar} illustrated.
In Ferret, a binary indicator $c_n^r$ within configuration $C$ denotes whether activation recomputation is enabled for the $n$-th worker.
When activation recomputation is enabled (\textit{i.e.}, $c_n^r=1$), an additional forward pass is executed prior to the backward pass, effectively managing memory consumption at the expense of increased computational load.

\textbf{T2. Gradient Accumulation:}
Gradient Accumulation allows multiple forward and backward passes to accumulate gradients before model updating, thereby decreasing the frequency of parameter updates, as Fig.~\ref{fig:prune-ga} depicted.
In Ferret, the parameter $c_{n,j}^a$ in configuration $C$ defaults to $1$, indicating the number of gradient accumulation steps before model updating for the $j$-th stage in the $n$-th worker.
By utilizing gradient accumulation, the $j$-th stage in the $n$-th worker only stores $(1 + \lceil (P - j - 1) / c_{n,j}^a \rceil)$, instead of $(P - j)$, models, thereby optimizing memory usage.

\textbf{T3. Back-propagation Omission:}
To further reduce memory usage, back-propagation omission skips all backward passes that depend on previous model parameters, as Fig.~\ref{fig:prune-bo} illustrated.
In Ferret, the parameter $c_{n,j}^o$ in configuration $C$ defaults to $0$, indicating the number of back-propagation omission steps for the $j$-th stage in the $n$-the worker.
This approach reduces memory overhead by eliminating the need to store multiple versions of models.

\textbf{T4. Worker Removal:}
Spawning $N$ workers increases the system throughput but also linearly increases the memory footprint.
When resources are highly constrained, the $n$-th worker can be shut down and removed to reduce the memory overhead by setting $c_n^d = -1$ in configuration $C$.

Finally, assume the initial data value of any data is $V_D$, and the retained value of the data when updating a subset of model parameters is proportional to the size of the subset model parameters, given $L$ and $C$, $\mathcal{R}$ and $\mathcal{M}$ of the fine-grained pipeline parallelism strategy can be respectively formulated as
{\small  
  \setlength{\abovedisplayskip}{6pt}
  \setlength{\belowdisplayskip}{\abovedisplayskip}
  \setlength{\abovedisplayshortskip}{0pt}
  \setlength{\belowdisplayshortskip}{3pt}
\begin{align}
    &\mathcal{R}_{F}^T = \sum_{\substack{n=1 \\ c_n^d \geq 0}}^{N-1} \sum_{i=0}^{P-1} \frac{|w_i|}{\sum_{j=0}^{P-1} (|w_j|)} \frac{1}{c_{n,i}^a} \sum_{j=0}^{c_{n,i}^a - 1} A_{i,j}, \mathrm{where}~A_{i,j} = \nonumber\\
    &\frac{e^{-c((P + j) t^f + (P-i+j) t^b + c_n^r (P-i+j) t^f)} V_D}{LCM(\{c_{n,k}^o + 1 | k \in [i,P-1]\}) (t^f + t^b + c_n^r t^f)}, \\
    &\mathcal{M}_{F} = \sum_{\substack{n=1 \\ c_n^d \geq 0}}^{N-1} \sum_{i=0}^{P-1} (1 + \lceil \frac{P - i - 1}{c_{n,i}^a} \rceil - c_{n,i}^o) (|w_i| + \nonumber\\
    &~~~~~~~~~~~~~~|a_i| - c_n^r \sum_{l=L_i + 1}^{L_{i+1}-1} |\hat{a}_l|),
\label{eq:ferret-svre}
\end{align}
}
where $LCM(\cdot)$ denotes the Least Common Multiple. 

\subsubsection{Iterative Gradient Compensation}

Since fine-grained pipeline parallelism is asynchronous, the model will be inevitably updated by stale gradients, leading to performance degradation.
Moreover, different pipeline stages of the model are updated by gradients with varying staleness.
To surmount the above challenges, Ferret firstly proposes to efficiently approximate $\nabla \mathcal{L}(D^{t-1}; \theta^{t})$ using $\nabla \mathcal{L}(D^{t-1}; \theta^{t-1})$ by a cost-effective approximator $A_{\mathcal{I}}(\cdot)$ based on Taylor series expansion and the Fisher information matrix.
Then, we extend this approximator to approximate $\nabla \mathcal{L}(D^{t-1}; \theta^{t+\tau-1})$ using $\nabla \mathcal{L}(D^{t-1}; \theta^{t-1})$ by iteratively applying $A_{\mathcal{I}}(\cdot)$.

\textbf{Gradients Compensation via Taylor Series Expansion:}
In prior work, $\nabla \mathcal{L}(D^{t-1}; \theta^{t})$ was naively set to $\nabla \mathcal{L}(D^{t-1}; \theta^{t-1})$~\cite{narayanan2019pipedream, narayanan2021memory} and can be regarded as a zero-order Taylor series expansion, leading to a high approximation error $||\nabla \mathcal{L}(D^{t-1}; \theta^{t}) - \nabla \mathcal{L}(D^{t-1}; \theta^{t-1})||^2$.
To reduce the approximation error, we expand $\nabla \mathcal{L}(D^{t-1}; \theta^{t})$ at $\theta^{t-1}$ by a first-order Taylor series expansion as follows:
{\small  
  \setlength{\abovedisplayskip}{6pt}
  \setlength{\belowdisplayskip}{\abovedisplayskip}
  \setlength{\abovedisplayshortskip}{0pt}
  \setlength{\belowdisplayshortskip}{3pt}
\begin{equation}
    \nabla \mathcal{L}(D^{t-1}; \theta^{t}) \approx \nabla \mathcal{L}(D^{t-1}; \theta^{t-1}) + \mathbb{H}(\mathcal{L}(D^{t-1}; \theta^{t-1})) \odot (\theta^{t} - \theta^{t-1}),
\label{eq:gt-taylor-expansion}
\end{equation}
}
where $\mathbb{H}(\cdot)$ denotes the Hessian matrix of $\cdot$.
Previous works have revealed that the Fisher information matrix (FIM) serves as an approximation of the Hessian matrix if $\mathcal{L}(\cdot)$ is a negative log-likelihood loss~\cite{friedman2001elements, papadopoulos2014information}.
Assuming $\theta^t$ gradually converges to its optimal value $\theta^*$ during training, we can achieve an unbiased estimation of $\mathbb{H}(\cdot)$ by:
{\small  
  \setlength{\abovedisplayskip}{6pt}
  \setlength{\belowdisplayskip}{\abovedisplayskip}
  \setlength{\abovedisplayshortskip}{0pt}
  \setlength{\belowdisplayshortskip}{3pt}
\begin{equation}
    \epsilon_t \triangleq \mathbb{E}_{D, \theta^*} || \mathcal{I}(\theta^t) - \mathbb{H}(\mathcal{L}(\cdot; \theta^{t})) || \rightarrow 0, t \rightarrow +\infty,
\end{equation}
}
where $\mathcal{I}(\theta)$ is the FIM.
To further mitigate space complexity, $\mathcal{I}(\theta)$ is approximated by its diagonal elements with a hyper-parameter $\lambda$ to control variance, \textit{i.e.},
{\small  
  \setlength{\abovedisplayskip}{6pt}
  \setlength{\belowdisplayskip}{\abovedisplayskip}
  \setlength{\abovedisplayshortskip}{0pt}
  \setlength{\belowdisplayshortskip}{3pt}
\begin{equation}
    \mathbb{H}(\mathcal{L}(\cdot; \theta^{t})) \approx \lambda \nabla \mathcal{L}(\cdot; \theta^{t-1}) \odot \nabla \mathcal{L}(\cdot; \theta^{t-1})^\top.
\label{eq:fisher-approximation}
\end{equation}
}
Incorporating Eq.\ref{eq:fisher-approximation} into Eq.\ref{eq:gt-taylor-expansion}, we obtain:
{\small  
  \setlength{\abovedisplayskip}{6pt}
  \setlength{\belowdisplayskip}{\abovedisplayskip}
  \setlength{\abovedisplayshortskip}{0pt}
  \setlength{\belowdisplayshortskip}{3pt}
\begin{align}
    &\nabla \mathcal{L}(D^{t-1}; \theta^{t}) \approx A_\mathcal{I}(\nabla \mathcal{L}(D^{t-1}; \theta^{t-1}), \theta^{t}, \theta^{t-1}) = \\
    &\nabla \mathcal{L}(D^{t-1}; \theta^{t-1}) + \lambda \nabla \mathcal{L}(D^{t-1}; \theta^{t-1}) \odot \nabla \mathcal{L}(D^{t-1}; \theta^{t-1})^\top \odot \Delta \theta,\nonumber
\label{eq:gt-taylor-expansion-approximation}
\end{align}
}
where $A_\mathcal{I}(\cdot)$ serves as the approximator to compensate $\cdot$.

\begin{figure}[t]
    \centering
    \includegraphics[width=\linewidth]{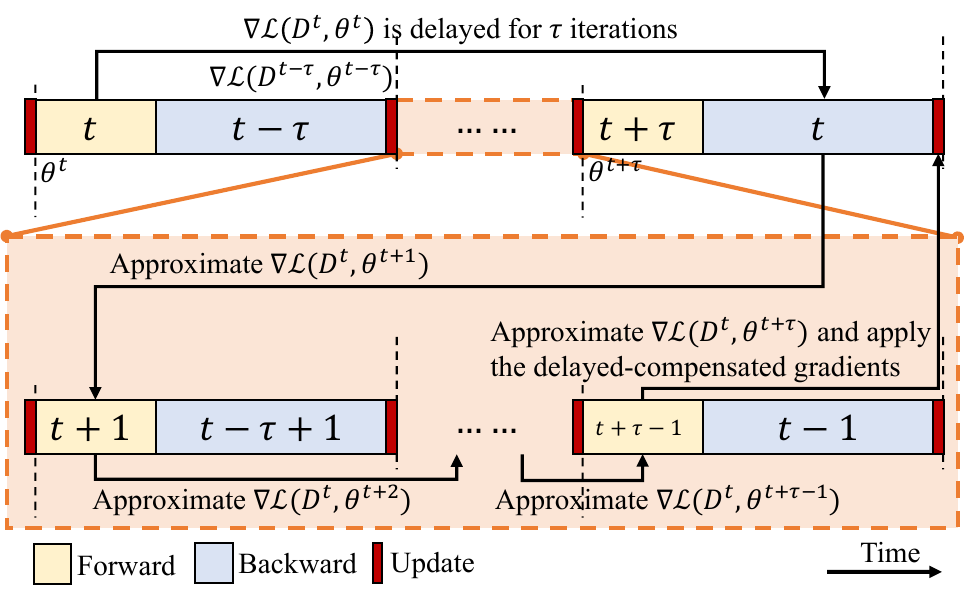}
    \caption{To adapt to different levels of staleness in fine-grained pipeline parallelism, $\nabla \mathcal{L}(D^t, \theta^{t+\tau})$ is iteratively approximated by $\nabla \mathcal{L}(D^t, \theta^{t})$.}
    \label{fig:iterative-gradients-compensation}
    \vspace{-15pt}
\end{figure}

\textbf{Iterative Compensation:}
More generally, to approximate $\nabla \mathcal{L}(D^{t-1}; \theta^{t+\tau-1})$ using $\nabla \mathcal{L}(D^{t-1}; \theta^{t-1})$, Ferret proposes an iterative application of $A_{\mathcal{I}}(\cdot)$, as depicted in Fig.~\ref{fig:iterative-gradients-compensation}. This iterative process is defined as follows:
{\small  
  \setlength{\abovedisplayskip}{6pt}
  \setlength{\belowdisplayskip}{\abovedisplayskip}
  \setlength{\abovedisplayshortskip}{0pt}
  \setlength{\belowdisplayshortskip}{3pt}
\begin{align}
    &\nabla \mathcal{L}(D^{t-1}; \theta^{t+\tau-1}) \approx A_\mathcal{I}(\nabla \mathcal{L}(D^{t-1}; \theta^{t+\tau-2}), \theta^{t+\tau-1}, \theta^{t+\tau-2}) \nonumber \\
    &\approx A_\mathcal{I}(\ldots A_\mathcal{I}(\nabla \mathcal{L}(D^{t-1}; \theta^{t-1}), \theta^{t}, \theta^{t-1}) \ldots, \theta^{t+\tau-1}, \theta^{t+\tau-2}).
\end{align}
}

However, this iterative process introduces a cascade of errors, wherein the approximation error $||g^{t} - A_\mathcal{I}(g^{t-1}, \theta^{t}, \theta^{t-1})||^2$ is propagated and amplified with each successive approximation.
This arises because each approximation depends on the output of the preceding one.

\begin{figure}[t]
    \centering
    \includegraphics[width=\linewidth]{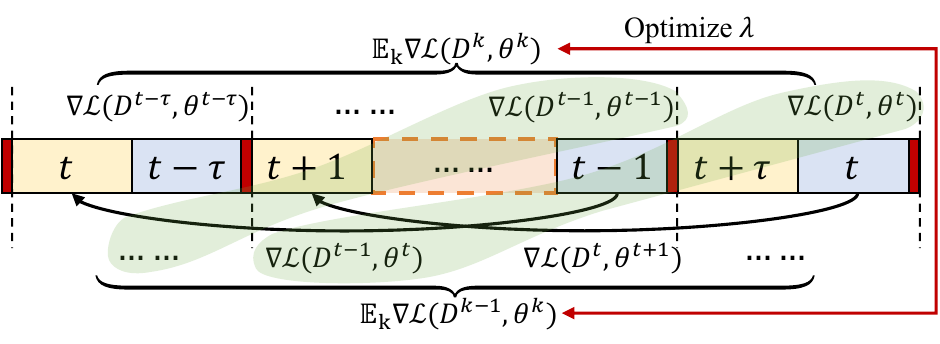}
    \caption{To further reduce approximation errors, we optimize $\lambda$ automatically by comparing historical approximations ($\nabla \mathcal{L}(D^t, \theta^t)$, etc.) and observations ($\nabla \mathcal{L}(D^{t-1}, \theta^t)$, etc.)}
    \label{fig:self-supervised-gc}
    \vspace{-15pt}
\end{figure}
To mitigate this problem, we propose to optimize $\lambda$ under the mild assumption that the distributions of $\mathbb{E}_k D^k$ and $\mathbb{E}_k D^{k+1}$ are similar, as illustrated in Fig.~\ref{fig:self-supervised-gc}.
Thus, the objective of minimizing the approximation error of iterative gradients compensation can be formulated as follows:
{\small  
  \setlength{\abovedisplayskip}{6pt}
  \setlength{\belowdisplayskip}{\abovedisplayskip}
  \setlength{\abovedisplayshortskip}{0pt}
  \setlength{\belowdisplayshortskip}{3pt}
\begin{align}
\label{eq:lambda-objective-v1}
    &\min_\lambda \mathbb{E}_k ||\nabla \mathcal{L}(D^{k}; \theta^{k}) - A_\mathcal{I}(\nabla \mathcal{L}(D^{k-1}, \theta^{k}, \theta^{k-1)}) ||^2 + \nu ||\lambda||^2 \nonumber \\
    &= \min_\lambda ||D - E - \lambda F||^2 + \nu ||\lambda||^2,  \\
    &\mathrm{where}~D=\mathbb{E}_k \nabla \mathcal{L}(D^{k}; \theta^{k}), ~E=\mathbb{E}_k \nabla \mathcal{L}(D^{k-1}; \theta^{k-1}), \nonumber \\
    &F=\mathbb{E}_k \nabla \mathcal{L}(D^{k-1}; \theta^{k-1}) \odot \nabla \mathcal{L}(D^{k-1}; \theta^{k-1})^\top \odot (\theta^k - \theta^{k-1}), \nonumber 
\end{align}
}
where $\nu ||\lambda||^2$ is an $\ell_2$ regularization term to constrain the solution of $\lambda$ for better stability. To reduce memory overhead, $D$ and $E$ can be approximated by Exponential Moving Average (EMA), \textit{i.e.},
{\small  
  \setlength{\abovedisplayskip}{6pt}
  \setlength{\belowdisplayskip}{\abovedisplayskip}
  \setlength{\abovedisplayshortskip}{0pt}
  \setlength{\belowdisplayshortskip}{3pt}
\begin{equation}
    \mathbb{E}_k \nabla \mathcal{L}(D^{k}; \theta^{k}) = \alpha \mathbb{E}_k \nabla \mathcal{L}(D^{k-1}; \theta^{k-1}) + (1 - \alpha) \nabla \mathcal{L}(D^{k-1}; \theta^{k-1}),
\end{equation}
}
where $\alpha$ is the EMA coefficient. Hence, we have
{\small  
  \setlength{\abovedisplayskip}{6pt}
  \setlength{\belowdisplayskip}{\abovedisplayskip}
  \setlength{\abovedisplayshortskip}{0pt}
  \setlength{\belowdisplayshortskip}{3pt}
\begin{equation}
    D - E = (1 - \alpha) (\nabla \mathcal{L}(D^{k-1}; \theta^{k-1}) - \mathbb{E}_k \nabla \mathcal{L}(D^{k-1}; \theta^{k-1})).
\end{equation}
}

\textbf{Convergence:}
Similar to the analyses in~\cite{zheng2017asynchronous}, our iterative gradient compensation algorithm yields convergence rates of $\mathcal{O}(V_1^2 \tau/T)$ and $\mathcal{O}(V_2/\sqrt{T})$ for convex and non-convex case, respectively.
Here, $V_1$ and $V_2$ represent the upper-bound of the $||\cdot||^2$ norm and the variance of the delay-compensated gradient $A_\mathcal{I}(\cdot)$, accordingly.
Compared to the work in~\cite{zheng2017asynchronous}, Ferret fixes $\tau$ to 1, and minimizes $V_1$ and $V_2$ by Eq.~\ref{eq:lambda-objective-v1}, boosting algorithm's robustness and accelerating the convergence of the model.

\textbf{Algorithm Design:}
The algorithm of Ferret's iterative gradient compensation is illustrated in Alg.~\ref{alg:iterative-gradients-compensation} in the appendix.
Since the maximum possible $\tau$ equals $(P-1)$, the time complexity of the algorithm is $\mathcal{O}(P-1)$, which is considered negligible during model training.
Moreover, since two additional variables, $v_r$ and $v_a$, are stored in memory for optimizing $\lambda$, the space complexity of this algorithm is $\mathcal{O}(2 \sum_{j=0}^{P-1} |w_i|)$.
However, by setting $\eta_\lambda = 0$, the optimization of $\lambda$ is effectively terminated, and $\lambda$ remains fixed at $\lambda^0$.
This adjustment allows for manual tuning of $\lambda$ and eliminates the need for $v_r$ and $v_a$, thereby increasing flexibility and avoiding additional memory overhead.

\subsection{Model Partitioning and Pipeline Planning}
The objective of model partitioning and pipeline planning is to find an optimal model partition scheme $L^*$ and its corresponding pipeline configuration $C^*$ that maximize $\mathcal{R}$ within a given memory constraint $M$, namely,
{\small  
  \setlength{\abovedisplayskip}{6pt}
  \setlength{\belowdisplayskip}{\abovedisplayskip}
  \setlength{\abovedisplayshortskip}{0pt}
  \setlength{\belowdisplayshortskip}{3pt}
\begin{equation}
    L^*, C^* = \arg \max_{L,C} \mathcal{R}_{F}^T~~\mathrm{s.t.}~~\mathcal{M}_{F} \leq M.
\label{eq:objective}
\end{equation}
}
This problem can be reformulated as a bi-level optimization problem, decomposing it into two interrelated sub-problems: (1) determining the optimal $C$ given a $L$, and (2) identifying the optimal $L$ based on the solution from (1):
{\small  
  \setlength{\abovedisplayskip}{6pt}
  \setlength{\belowdisplayskip}{\abovedisplayskip}
  \setlength{\abovedisplayshortskip}{0pt}
  \setlength{\belowdisplayshortskip}{3pt}
\begin{align}
    L^* &= \arg \max_L \{\mathcal{R}_{F}^T | C_L^*\} \nonumber\\
    ~~s.t.~~C_L^*&=\arg \max_C \{\mathcal{R}_{F}^T | L\}, \mathcal{M}_{F} \leq M.
\end{align}
}

\subsubsection{Iterative Configuration Search (Sub-problem 1)}
\label{sec:ics}

Given a model partition scheme, the objective of sub-problem (1) is to solve
{\small  
  \setlength{\abovedisplayskip}{6pt}
  \setlength{\belowdisplayskip}{\abovedisplayskip}
  \setlength{\abovedisplayshortskip}{0pt}
  \setlength{\belowdisplayshortskip}{3pt}
\begin{equation}
    C^* = \arg \max_{C} \{\mathcal{R}_{F}^T | L\}~~\mathrm{s.t.}~~\mathcal{M}_{F} \leq M.
\label{eq:objective-given-L}
\end{equation}
}
With more than $2^{N(P+1)}$ potential combinations for $C$, a brute-force enumeration of $C$ is impractical.
Observing that $\mathrm{d}\mathcal{M}_F / \mathrm{d}\max_{C} \{\mathcal{R}_F^T | L\} \geq 0$, we employ an iterative algorithm to determine the optimal $C$ that maximize $\mathcal{R}_{F}^T$ while ensuring $\mathcal{M}_{F}$ remains within the memory budget.
Specifically, to prevent memory over-consumption, we progressively deploy \textbf{T1}-\textbf{T4} as follows to balance $\mathcal{R}_{F}^T$ and $\mathcal{M}_F$.

\textbf{S1. Deploy T1 for all workers: } 
By setting $c_n^r=1$ for all workers, the data processing time increases.
Specifically, for the $n$-th worker, setting $c_n^r=1$ will respectively reduce $\mathcal{R}_{F}^T$ and $\mathcal{M}_{F}$ by Eq.~\ref{eq:dsvre-ar} in the appendix.

\textbf{S2. Deploy T2 for the $j$-th stage in the $n$-th worker: }
If $c_{n,j}^o=0$, increasing $c_{n,j}^a$ by $\Delta c_{n,j}^a = \lceil \frac{P - j - 1}{\lceil (P - j - 1) / c_{n,j}^a \rceil - 1} \rceil - c_{n,j}^a$ will lead to a reduced frequency of model parameter updates.
Here, the value of $\Delta c_{n,j}^a$ is determined to prevent $\Delta_{c_{n,j}^a \rightarrow c_{n,j}^a + 1} \mathcal{M}_{F} = 0$ due to the ceiling function.
Consequently, $\mathcal{R}_{F}^T$ and $\mathcal{M}_{F}$ will be respectively decreased by Eq.~\ref{eq:dsvre-ga} in the appendix.

\textbf{S3. Deploy T3 For the $j$-th stage in the $n$-th worker: } 
If $\Delta c_{n,j}^a = +\infty$, setting $c_{n,j}^a = 1$ and $c_{n,j}^o=P - 1 - j$ will completely eliminate the need for the $j$-th stage in the $n$-th worker to store additional model parameters by bypassing any backward pass that requires previous model parameters.
Consequently, $\mathcal{R}_{F}^T$ and $\mathcal{M}_{F}$ will be respectively reduced by Eq.~\ref{eq:dsvre-bpo} in the appendix.

\textbf{S4. Deploy T4 for the $n$-th worker: }
If $c_{n,j}^o \neq 0$ for all $j \in [0, p-1)$, removing the $n$-th worker will lead to a decrease in $\mathcal{R}_{F}^T$ and $\mathcal{M}_{F}$ by Eq.~\ref{eq:dsvre-d} in the appendix.

\textbf{Algorithm Design:}
The algorithm of the proposed searching is illustrated in Alg.~\ref{alg:iter-conf-search} in the appendix.
Overall, the time complexity of this algorithm is $\mathcal{O}(NP^2)$, and it will be executed only once before fine-grained pipeline parallelism begins.

\subsubsection{Brute-force Planning (Sub-problem 2)}

In Ferret, $L$ is determined by first establishing an upper bound on the time consumed for each stage ($t^c$), and then solving the following optimization problem:
{\small  
  \setlength{\abovedisplayskip}{6pt}
  \setlength{\belowdisplayskip}{\abovedisplayskip}
  \setlength{\abovedisplayshortskip}{0pt}
  \setlength{\belowdisplayshortskip}{3pt}
\begin{equation}
    L = \arg \min_L \{P\}~~s.t.~~t^f + t^b \leq t^c.
\end{equation}
}
Namely, minimizing the number of pipeline stages while ensuring the time consumed for each stage is bounded.
Since the layers in a stage must be consecutive, this problem can be solved in linear time by iteratively grouping consecutive layers into a stage until no additional adjacent layer can be grouped.
Therefore, the solution space for $L$ is not extensive, being limited to $(\hat{L}^2 - \hat{L}) / 2$ at worst.
Thus, to solve sub-problem (2), we can simply enumerate all possible model partition schemes, feeding them into Alg.~\ref{alg:iter-conf-search} in the appendix to obtain the global optimum $L^*$.

\textbf{Algorithm Design:}
The algorithm of the proposed planning is illustrated in Alg.~\ref{alg:brute-force-planning} in the appendix.
The time complexity of this algorithm is $\mathcal{O}(\hat{L}^3)$.
Nevertheless, the algorithm will be executed only once before fine-grained pipeline parallelism begins.

\section{Experiments}
\label{sec:exp}

In this section, we seek answers to the following questions. 
(1) How does Ferret boost online accuracy? (Sec.~\ref{sec:overall-comparisons})
(2) How does Ferret mitigate catastrophic forgetting. (Sec.~\ref{sec:overall-comparisons})
(3) How does our fine-grained pipeline parallelism perform? (Sec.~\ref{sec:pp-comparisons})
(4) What are influences of different pipeline configurations? (Sec.~\ref{sec:pp-comparisons})
(5) How does our iterative gradient compensation algorithm perform?(Sec.~\ref{sec:gc-comparisons})

\subsection{Evaluation Setup}

\begin{figure*}
  \begin{minipage}{0.75\textwidth}
    \centering
    \captionof{table}{Online Accuracy Gain per unit of Memory ($agm_{\mathcal{B}}(\mathcal{A}, T)$) of different algorithms, where $\mathcal{B}$ is the 1-Skip. "M-", "M", "M+" refer to the ferret method with minimal, medium and maximal memory footprint, respectively.}
    \resizebox{1.0\linewidth}{!}{%
      \begin{tabular}{l|r|rrrr|rrr}
      \toprule
      Setting & \cellcolor{gray!20}Oracle  & 1-Skip  & Random-$N$ & Last-$N$ & Camel & Ferret$_{\mathrm{M-}}$ & Ferret$_{\mathrm{M}}$ & Ferret$_{\mathrm{M+}}$ \\
      \midrule
      MNIST/MNISTNet & \cellcolor{gray!20}27.32$_{\pm0.71}$ & 0$_{\pm0}$ & -0.43$_{\pm0.6}$ & -0.26$_{\pm0.14}$ & -0.71$_{\pm0.32}$ & 5.31$_{\pm0.7}$ & \underline{16.26}$_{\pm0.37}$ & \textbf{26.34}$_{\pm0.7}$ \\
      FMNIST/MNISTNet & \cellcolor{gray!20}19.35$_{\pm0.99}$ & 0$_{\pm0}$ & -0.31$_{\pm0.47}$ & -0.25$_{\pm0.5}$ & -0.6$_{\pm0.4}$ & 5.93$_{\pm0.81}$ & \underline{12.69}$_{\pm0.81}$ & \textbf{18.37}$_{\pm1.01}$ \\
      EMNIST/MNISTNet & \cellcolor{gray!20}13$_{\pm0.48}$ & 0$_{\pm0}$ & 1.94$_{\pm0.04}$ & 2.02$_{\pm0.03}$ & 1.55$_{\pm0.1}$ & 4.19$_{\pm0.17}$ & \underline{8.8}$_{\pm0.4}$ & \textbf{12.09}$_{\pm0.47}$ \\
      CIFAR10/ConvNet & \cellcolor{gray!20}10.57$_{\pm0.09}$ & 0$_{\pm0}$ & 4.71$_{\pm0.05}$ & 4.78$_{\pm0.03}$ & 4.7$_{\pm0.05}$ & 3.21$_{\pm0.16}$ & \underline{6.21}$_{\pm0.15}$ & \textbf{9.44}$_{\pm0.12}$ \\
      CIFAR100/ConvNet & \cellcolor{gray!20}5.24$_{\pm0.01}$ & 0$_{\pm0}$ & 0.78$_{\pm0.07}$ & 0.83$_{\pm0.06}$ & 0.75$_{\pm0.08}$ & 1.58$_{\pm0.04}$ & \underline{2.6}$_{\pm0.03}$ & \textbf{4.39}$_{\pm0.05}$ \\
      SVHN/ConvNet & \cellcolor{gray!20}15.41$_{\pm0.23}$ & 0$_{\pm0}$ & 7.04$_{\pm0.08}$ & 7.24$_{\pm0.11}$ & 7.39$_{\pm0.09}$ & 5$_{\pm0.1}$ & \underline{11.52}$_{\pm0.23}$ & \textbf{14.34}$_{\pm0.31}$ \\
      TinyImagenet/ConvNet & \cellcolor{gray!20}2.13$_{\pm0.07}$ & 0$_{\pm0}$ & -0.22$_{\pm0.04}$ & -0.2$_{\pm0.03}$ & -0.21$_{\pm0.05}$ & 0.48$_{\pm0.03}$ & \underline{0.54}$_{\pm0.03}$ & \textbf{1.19}$_{\pm0.07}$ \\
      CORe50/ConvNet & \cellcolor{gray!20}26.01$_{\pm0.42}$ & 0$_{\pm0}$ & 12.13$_{\pm0.42}$ & 12.27$_{\pm0.44}$ & 11.07$_{\pm0.48}$ & 9.08$_{\pm0.45}$ & \underline{17.95}$_{\pm0.45}$ & \textbf{24.49}$_{\pm0.43}$ \\
      CORe50-iid/ConvNet & \cellcolor{gray!20}19.24$_{\pm2.9}$ & 0$_{\pm0}$ & 2.87$_{\pm5.71}$ & 5.74$_{\pm2.8}$ & 5.21$_{\pm2.49}$ & 3.55$_{\pm2.77}$ & \underline{10.74}$_{\pm2.76}$ & \textbf{17.96}$_{\pm2.88}$ \\
      SplitMNIST/MNISTNet & \cellcolor{gray!20}18.21$_{\pm0.76}$ & 0$_{\pm0}$ & 2.34$_{\pm0.43}$ & 2.37$_{\pm0.63}$ & 3.3$_{\pm0.48}$ & 6.11$_{\pm0.84}$ & \underline{14.55}$_{\pm0.55}$ & \textbf{17.05}$_{\pm0.72}$ \\
      SplitFMNIST/MNISTNet & \cellcolor{gray!20}11.32$_{\pm1.47}$ & 0$_{\pm0}$ & 1.53$_{\pm0.47}$ & 1.49$_{\pm0.42}$ & 1.96$_{\pm0.39}$ & 5.43$_{\pm0.56}$ & \underline{9.37}$_{\pm1.35}$ & \textbf{10.29}$_{\pm1.47}$ \\
      SplitCIFAR10/ConvNet & \cellcolor{gray!20}7.49$_{\pm0.12}$ & 0$_{\pm0}$ & 3.05$_{\pm0.11}$ & 3.11$_{\pm0.11}$ & 3.12$_{\pm0.09}$ & 2.91$_{\pm0.19}$ & \underline{4.84}$_{\pm0.2}$ & \textbf{6.19}$_{\pm0.07}$ \\
      SplitCIFAR100/ConvNet & \cellcolor{gray!20}10.51$_{\pm0.15}$ & 0$_{\pm0}$ & 2.81$_{\pm0.07}$ & 2.86$_{\pm0.05}$ & 2.74$_{\pm0.13}$ & 3.54$_{\pm0.03}$ & \underline{6.13}$_{\pm0.13}$ & \textbf{9.61}$_{\pm0.04}$ \\
      SplitSVHN/ConvNet & \cellcolor{gray!20}6.49$_{\pm0.33}$ & 0$_{\pm0}$ & 2.9$_{\pm0.19}$ & 2.91$_{\pm0.21}$ & 2.89$_{\pm0.21}$ & 2.76$_{\pm0.16}$ & \underline{5}$_{\pm0.28}$ & \textbf{5.38}$_{\pm0.35}$ \\
      SplitTinyImagenet/ConvNet & \cellcolor{gray!20}2.14$_{\pm0.1}$ & 0$_{\pm0}$ & -0.24$_{\pm0.03}$ & -0.21$_{\pm0.02}$ & -0.26$_{\pm0.01}$ & 0.47$_{\pm0.01}$ & \underline{0.62}$_{\pm0.01}$ & \textbf{1.19}$_{\pm0.06}$ \\
      CLEAR10/ResNet & \cellcolor{gray!20}10.37$_{\pm0.06}$ & 0$_{\pm0}$ & 7.84$_{\pm0.07}$ & \underline{7.93}$_{\pm0.06}$ & -2.9$_{\pm10.55}$ & 2.44$_{\pm0.06}$ & 7.71$_{\pm0.06}$ & \textbf{9.26}$_{\pm0.08}$ \\
      CLEAR10/MobileNet & \cellcolor{gray!20}20.36$_{\pm0.2}$ & 0$_{\pm0}$ & 11.8$_{\pm0.22}$ & 12$_{\pm0.14}$ & 11.85$_{\pm0.07}$ & -1.77$_{\pm0.15}$ & \underline{14.68}$_{\pm0.5}$ & \textbf{18.51}$_{\pm0.35}$ \\
      CLEAR100/ResNet & \cellcolor{gray!20}21.71$_{\pm0.43}$ & 0$_{\pm0}$ & 15.19$_{\pm0.49}$ & 15.36$_{\pm0.46}$ & 14.39$_{\pm0.46}$ & 7.51$_{\pm0.44}$ & \underline{15.53}$_{\pm0.35}$ & \textbf{20.84}$_{\pm0.57}$ \\
      CLEAR100/MobileNet & \cellcolor{gray!20}23.51$_{\pm1.03}$ & 0$_{\pm0}$ & 9.16$_{\pm0.28}$ & 9.39$_{\pm0.15}$ & 8.72$_{\pm0.06}$ & 1.05$_{\pm0.13}$ & \underline{15.8}$_{\pm0.39}$ & \textbf{22.11}$_{\pm0.59}$ \\
      Covertype/MLP & \cellcolor{gray!20}7.66$_{\pm0.27}$ & 0$_{\pm0}$ & -1.33$_{\pm0.3}$ & -1.3$_{\pm0.31}$ & -1.34$_{\pm0.29}$ & 0.74$_{\pm0.21}$ & \underline{1.61}$_{\pm0.29}$ & \textbf{3.38}$_{\pm0.42}$ \\
      \bottomrule
      \end{tabular}%
      }
    \label{tab:overall-table}%
  \end{minipage}%
  \hfill
  \begin{minipage}{0.24\textwidth}
    \centering
    \captionsetup[subfloat]{labelfont=scriptsize,textfont=scriptsize}
    \subfloat[EMNIST/MNISTNet]{\includegraphics[width=0.7\linewidth]{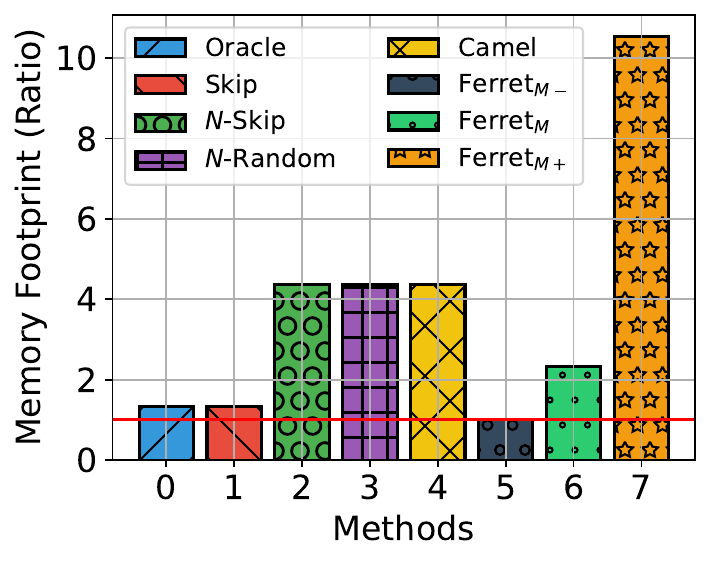}}
    \\
    \subfloat[CIFAR100/ConvNet]{\includegraphics[width=0.7\linewidth]{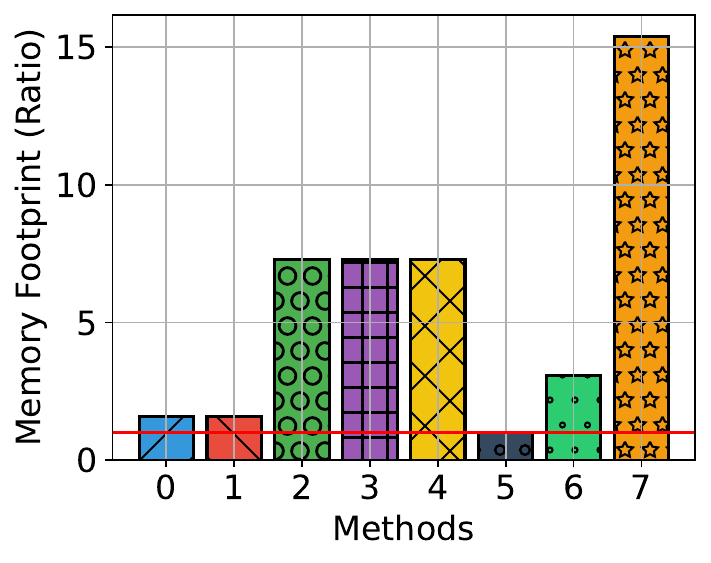}}
    \captionof{figure}{Consumed memory of different stream learning algorithms. Ferret achieves rapid adaptation across varying memory constraints.}
    \label{fig:overall-memory-comparison}
  \end{minipage}
\end{figure*}

\begin{figure*}
  \begin{minipage}{0.37\textwidth}
    \centering
    \captionsetup[subfloat]{labelfont=scriptsize,textfont=scriptsize}
    \subfloat[EMNIST/MNISTNet]{\includegraphics[width=0.5\linewidth]{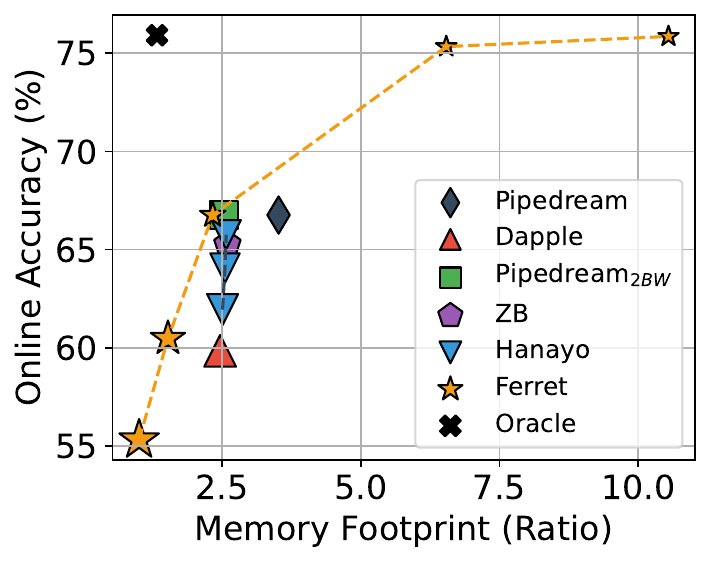}}
    \subfloat[CORe50/ConvNet]{\includegraphics[width=0.5\linewidth]{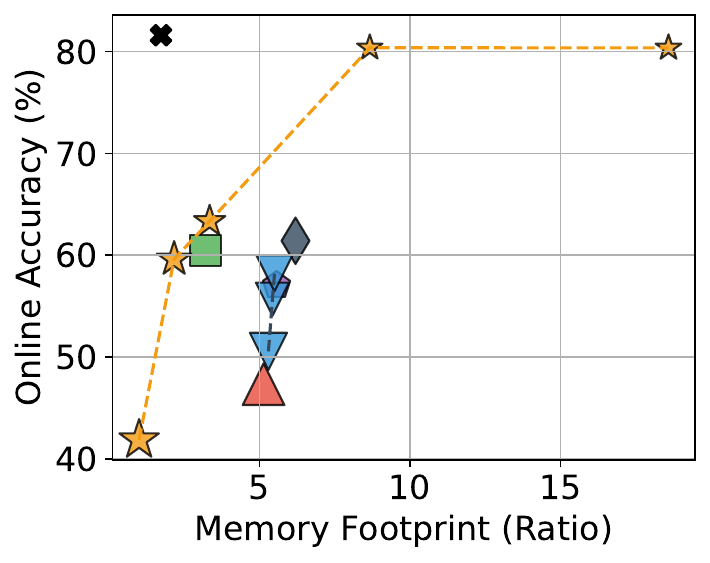}}
    \captionof{figure}{Relationships between online accuracy and memory consumption of different pipeline parallelism strategies, the marker size represents the standard errors of means.}
    \label{fig:pp-memory-comparison}
  \end{minipage}%
  \hfill
  \begin{minipage}{0.62\textwidth}
    \centering
    \captionof{table}{Online Accuracy Gain per unit of Memory ($agm_{\mathcal{B}}(\mathcal{A}, T)$) and Test Accuracy Gain per unit of Memory ($tagm_{\mathcal{B}}(\mathcal{A})$) of different integrated OCL algorithms on CORe50/ConvNet, where $\mathcal{B}$ is the 1-Skip. Camel has its dedicated component to mitigate catastrophic forgetting and cannot be integrated with various OCL algorithm.}
    \resizebox{1.0\linewidth}{!}{%
    \begin{tabular}{rl|l|llll|lll}
    \toprule
          & Metric & \cellcolor{gray!20}Oracle  & 1-Skip  & Random-$N$ & Last-$N$ & Camel & Ferret$_{\mathrm{M-}}$ & Ferret$_{\mathrm{M}}$ & Ferret$_{\mathrm{M+}}$ \\
    \midrule
    \multicolumn{1}{l}{Vanilla} & $agm$     & \cellcolor{gray!20}26.01$_{\pm0.42}$ & 0$_{\pm0}$ & 12.13$_{\pm0.42}$ & 12.27$_{\pm0.44}$ & 11.07$_{\pm0.48}$ & 9.08$_{\pm0.45}$ & \underline{17.21}$_{\pm0.45}$ & \textbf{24.82}$_{\pm0.43}$ \\
          & $tagm$     & \cellcolor{gray!20}2.36$_{\pm0.64}$ & 0$_{\pm0}$ & 1.08$_{\pm0.62}$ & 0.97$_{\pm0.39}$ & \underline{1.48}$_{\pm0.42}$ & 1.01$_{\pm0.45}$ & 1.07$_{\pm0.49}$ & \textbf{1.73}$_{\pm0.53}$ \\
    \multicolumn{1}{l}{ER~\cite{chaudhry2019continual}} & $agm$     & \cellcolor{gray!20}24.03$_{\pm0.26}$ & 0$_{\pm0}$ & 7.84$_{\pm0.17}$ & 8.11$_{\pm0.29}$ & -      & 7.12$_{\pm0.09}$ & \underline{16.09}$_{\pm0.16}$ & \textbf{23.5}$_{\pm0.25}$ \\
          & $tagm$     & \cellcolor{gray!20}4.18$_{\pm0.43}$ & 0$_{\pm0}$ & 1.94$_{\pm0.26}$ & 2.34$_{\pm0.29}$ & -    & 0.82$_{\pm0.29}$ & \underline{3.1}$_{\pm0.25}$ & \textbf{4.06}$_{\pm0.36}$ \\
    \multicolumn{1}{l}{MIR~\cite{aljundi2019online}} & $agm$     & \cellcolor{gray!20}24.03$_{\pm0.26}$ & 0$_{\pm0}$ & 7.82$_{\pm0.21}$ & 8.06$_{\pm0.22}$ & -      & 7.12$_{\pm0.09}$ & \underline{16.09}$_{\pm0.15}$ & \textbf{23.5}$_{\pm0.25}$ \\
          & $tagm$     & \cellcolor{gray!20}4.18$_{\pm0.43}$ & 0$_{\pm0}$ & 2.1$_{\pm0.15}$ & 2.2$_{\pm0.14}$ & -      & 0.82$_{\pm0.29}$ & \underline{3.1}$_{\pm0.25}$ & \textbf{4.06}$_{\pm0.36}$ \\
    \multicolumn{1}{l}{LwF~\cite{li2017learning}} & $agm$     & \cellcolor{gray!20}26.02$_{\pm0.42}$ & 0$_{\pm0}$ & 12.25$_{\pm0.42}$ & 12.4$_{\pm0.45}$ & -      & 9.02$_{\pm0.4}$ & \underline{17.96}$_{\pm0.46}$ & \textbf{24.67}$_{\pm0.4}$ \\
          & $tagm$     & \cellcolor{gray!20}2.36$_{\pm0.64}$ & 0$_{\pm0}$ & 1.2$_{\pm0.62}$ & 1.09$_{\pm0.39}$ & -      & 0.91$_{\pm0.44}$ & \underline{1.88}$_{\pm0.49}$ & \textbf{1.54}$_{\pm0.53}$ \\
    \multicolumn{1}{l}{MAS~\cite{aljundi2018memory}} & $agm$     & \cellcolor{gray!20}25.86$_{\pm0.25}$ & 0$_{\pm0}$ & 12.04$_{\pm0.23}$ & 12.23$_{\pm0.22}$ & -      & 8.7$_{\pm0.17}$ & \underline{17.79}$_{\pm0.22}$ & \textbf{24.46}$_{\pm0.23}$ \\
          & $tagm$     & \cellcolor{gray!20}2.7$_{\pm0.23}$ & 0$_{\pm0}$ & 0.81$_{\pm0.41}$ & 0.91$_{\pm0.24}$ & -      & 0.59$_{\pm0.23}$ & \underline{1.66}$_{\pm0.18}$ & \textbf{1.69}$_{\pm0.21}$ \\
    \bottomrule
    \end{tabular}%
    }
    \label{tab:underlying-A}%
  \end{minipage}
\end{figure*}

\textbf{Datasets and models: }
Following the conventions of the community~\cite{li2022camel}, 18 image classification datasets, including MNIST~\cite{deng2012mnist}, FMNIST~\cite{xiao2017fashion}, CIFAR10~\cite{krizhevsky2009learning}, CIFAR100~\cite{krizhevsky2009learning}, SVHN~\cite{netzer2011reading}, Tiny-ImageNet~\cite{le2015tiny}, CORe50~\cite{lomonaco2017core50}, CORe50-iid, Split-MNIST, Split-FMNIST, Split-CIFAR10, Split-CIFAR100, Split-SVHN, Split-Tiny-ImageNet, Covertype~\cite{misc_covertype_31}, CLEAR10~\cite{lin2021clear}, CLEAR100~\cite{lin2021clear}, are used in our experiments.
More details about the datasets can be found in the appendix.
To cover both simple and complicated learning problems, five models including Multi-Layer Perceptron (MLP), MNISTNet, ConvNet, ResNet-18~\cite{he2016deep} and MobileNet~\cite{howard2017mobilenets} are used in the experiments. 
Note that ResNet-18 and MobileNet are pretrained on the ImageNet-1K dataset~\cite{deng2009imagenet}.

\begin{wrapfigure}{r}[0pt]{0.4\linewidth}
  \centering
  \includegraphics[width=\linewidth]{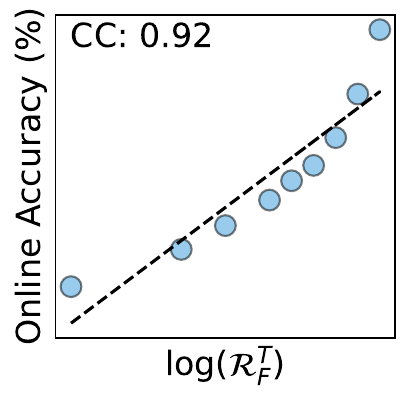}
  \vspace{-25pt}
  \caption{Relation between $oacc$ and $\log (\mathcal{R}_F^T)$}
  \label{fig:r-to-acc-relation}
\end{wrapfigure}
\textbf{Compared methods: }
For question (1): \textbf{Oracle}, \textbf{1-Skip}~\cite{ghunaim2023real}, \textbf{Random-$N$ $B$-Skip}, \textbf{Last-$N$ $B$-Skip} and Camel~\cite{li2022camel}.
Here, Oracle is an ideal method that sequentially processes every streaming data without any delay.
$B$-Skip and Camel selects a subset from the latest $B$ unprocessed data using Random-$N$, Last-$N$ and Coreset sampler, respectively.
For question (2): \textbf{Vanilla}, \textbf{ER}~\cite{chaudhry2019continual}, \textbf{MIR}~\cite{aljundi2019online}, \textbf{LwF}~\cite{li2017learning}, \textbf{MAS}~\cite{aljundi2018memory}.
For question (3) and (4): \textbf{DAPPLE}~\cite{fan2021dapple}, \textbf{Pipedream}~\cite{narayanan2019pipedream}, \textbf{Pipedream$_{2BW}$}~\cite{narayanan2021memory}, \textbf{Zero-Bubble}~\cite{qi2023zero} and \textbf{Hanayo}~\cite{liu2023hanayo}.
For question (5): \textbf{None}, \textbf{Step-Aware}~\cite{hardy2017distributed, jiang2017heterogeneity}, \textbf{Gap-Aware}~\cite{barkai2019gap}, \textbf{Fisher}~\cite{chen2022sapipe}, and \textbf{Iter-Fisher} (\textit{i.e.}, iterative gradient compensation).

\textbf{Evaluation metrics: } 
To measure catastrophic forgetting while accounting for memory footprint, Test Accuracy~\cite{li2022camel, ghunaim2023real} Gain per unit of Memory\footnote{Results in the appendix show standard test accuracy and online accuracy but ignore memory consumption during training.\label{ftn:startard-exps}} (the higher the better) is defined as
{\small  
  \setlength{\abovedisplayskip}{6pt}
  \setlength{\belowdisplayskip}{\abovedisplayskip}
  \setlength{\abovedisplayshortskip}{0pt}
  \setlength{\belowdisplayshortskip}{3pt}
\begin{equation}
    tagm_{\mathcal{B}}(\mathcal{A}, t) = \log(\frac{\exp({tacc_\mathcal{A}(t) - tacc_\mathcal{B}(t)})}{\mathcal{M}_\mathcal{A} / \mathcal{M}_\mathcal{B}}),
\label{eq:metric-tagm}
\end{equation}
}
where $tacc$ computes the test accuracy and $\mathcal{B}$ is the baseline method for comparison.
Similarly, as Fig.~\ref{fig:r-to-acc-relation} shows online accuracy can be used to estimate $\mathcal{R}_F^T$~\cite{cai2021online}, Online Accuracy Gain per unit of Memory\textsuperscript{\ref{ftn:startard-exps}} (the higher the better) is defined as
{\small  
  \setlength{\abovedisplayskip}{6pt}
  \setlength{\belowdisplayskip}{\abovedisplayskip}
  \setlength{\abovedisplayshortskip}{0pt}
  \setlength{\belowdisplayshortskip}{3pt}
\begin{equation}
    agm_{\mathcal{B}}(\mathcal{A}, t) = \log(\frac{\exp({oacc_\mathcal{A}(t) - oacc_\mathcal{B}(t)})}{\mathcal{M}_\mathcal{A} / \mathcal{M}_\mathcal{B}}).
\label{eq:metric-agm}
\end{equation}
}

We evaluate three versions of Ferret under different memory constraints: Ferret$_{\mathrm{M-}}$ (minimal), Ferret$_{\mathrm{M}}$ (the same memory constraint as Pipedream$_{2BW}$), and Ferret$_{\mathrm{M+}}$ (no constraint).
Without clarification, each experiment is independently repeated three times to obtain the final results. 
In all tables, the best and second-best performance are highlighted by \textbf{bold} and \underline{underline}, respectively.
More details about the evaluation setup can be found in Sec.~\ref{sec:implementation}.

\begin{table*}[htbp]
  \begin{minipage}{.527\linewidth}
    \centering
    \caption{Online Accuracy Gain per unit of Memory ($agm_{\mathcal{B}}(\mathcal{A}, T)$) of different \colorbox{cyan!20}{pipeline parallelism strategies}, where $\mathcal{B}$ is the DAPPLE. Note that "1W", "2W" and "3W" refer to 1, 2 and 3 wave(s) for the Hanayo algorithm, and no gradients compensation is applied to all asynchronous pipeline parallelism strategies for fair comparisons}
  \resizebox{\linewidth}{!}{%
    \begin{tabular}{l|rrrrr|rrr|}
    \toprule
    \multirow{2}[4]{*}{Setting} & \multicolumn{5}{c}{\cellcolor{cyan!20}Synchronous PP}           & \multicolumn{3}{c}{\cellcolor{cyan!20}Asynchronous PP} \\
\cmidrule{2-9}          & \cellcolor{cyan!20}DAPPLE & \cellcolor{cyan!20}ZB    & \cellcolor{cyan!20}Hanayo$_{\mathrm{1W}}$ & \cellcolor{cyan!20}Hanayo$_{\mathrm{2W}}$ & \cellcolor{cyan!20}Hanayo$_{\mathrm{3W}}$ & \cellcolor{cyan!20}Pipedream & \cellcolor{cyan!20}Pipedream$_{\mathrm{2BW}}$ & \cellcolor{cyan!20}Ferret$_{\mathrm{M}}$ \\
    \midrule
    MNIST/MnNet & 0$_{\pm0}$ & 6.79$_{\pm0.4}$ & 2.44$_{\pm0.3}$ & 5$_{\pm0.16}$ & 7.12$_{\pm0.38}$ & 8.16$_{\pm0.35}$ & \underline{8.23}$_{\pm0.39}$ & \textbf{8.35}$_{\pm0.35}$ \\
FMNIST/MnNet & 0$_{\pm0}$ & 4.06$_{\pm0.36}$ & 1.52$_{\pm0.45}$ & 2.8$_{\pm0.65}$ & 4.26$_{\pm0.64}$ & 5.29$_{\pm0.53}$ & \underline{5.36}$_{\pm0.54}$ & \textbf{5.48}$_{\pm0.53}$ \\
EMNIST/MnNet & 0$_{\pm0}$ & 2.33$_{\pm0.08}$ & 0.9$_{\pm0.02}$ & 1.81$_{\pm0.05}$ & 2.55$_{\pm0.04}$ & 2.84$_{\pm0.09}$ & \underline{2.99}$_{\pm0.07}$ & \textbf{3.02}$_{\pm0.09}$ \\
C10/CNet & 0$_{\pm0}$ & 1.76$_{\pm0.08}$ & 0.96$_{\pm0.14}$ & 1.51$_{\pm0.04}$ & 1.93$_{\pm0.12}$ & 2.53$_{\pm0.04}$ & \underline{2.78}$_{\pm0.06}$ & \textbf{3.05}$_{\pm0.15}$ \\
C100/CNet & 0$_{\pm0}$ & 0.71$_{\pm0.04}$ & 0.05$_{\pm0.05}$ & 0.56$_{\pm0.06}$ & 0.74$_{\pm0.06}$ & 0.87$_{\pm0.09}$ & \underline{1.11}$_{\pm0.06}$ & \textbf{1.72}$_{\pm0.01}$ \\
SVHN/CNet & 0$_{\pm0}$ & 2.13$_{\pm0.32}$ & 0.36$_{\pm0.15}$ & 1.52$_{\pm0.23}$ & 2.21$_{\pm0.26}$ & 3.3$_{\pm0.24}$ & \underline{3.32}$_{\pm0.19}$ & \textbf{3.61}$_{\pm0.16}$ \\
TinyI/CNet & 0$_{\pm0}$ & 0.18$_{\pm0.02}$ & 0.03$_{\pm0.01}$ & 0.19$_{\pm0.02}$ & 0.19$_{\pm0.04}$ & 0.26$_{\pm0.01}$ & \textbf{0.52}$_{\pm0.03}$ & \underline{0.5}$_{\pm0.06}$ \\
CORe50/CNet & 0$_{\pm0}$ & 4.18$_{\pm0.23}$ & 1.38$_{\pm0.12}$ & 3.6$_{\pm0.16}$ & 4.69$_{\pm0.11}$ & \underline{6.03}$_{\pm0.17}$ & 5.91$_{\pm0.22}$ & \textbf{7.13}$_{\pm0.18}$ \\
CORe50-iid/CNet & 0$_{\pm0}$ & 3.58$_{\pm0.02}$ & 1.19$_{\pm0.18}$ & 2.94$_{\pm0.02}$ & 3.74$_{\pm0.07}$ & 5.07$_{\pm0.13}$ & \underline{5.24}$_{\pm0.07}$ & \textbf{6.18}$_{\pm0.05}$ \\
S-MNIST/MnNet & 0$_{\pm0}$ & 2.94$_{\pm0.29}$ & 1.33$_{\pm0.26}$ & 2.97$_{\pm0.21}$ & 3.69$_{\pm0.23}$ & \underline{4.3}$_{\pm0.29}$ & 4.11$_{\pm0.33}$ & \textbf{4.47}$_{\pm0.29}$ \\
S-FMNIST/MnNet &0$_{\pm0}$ & 1.56$_{\pm0.29}$ & 0.91$_{\pm0.16}$ & 1.48$_{\pm0.2}$ & 1.89$_{\pm0.31}$ &  2.06$_{\pm0.28}$ & \underline{2.09}$_{\pm0.27}$ & \textbf{2.24}$_{\pm0.28}$ \\
S-C10/CNet & 0$_{\pm0}$ & 0.96$_{\pm0.14}$ & 0.44$_{\pm0.13}$ & 1.19$_{\pm0.03}$ & 1.42$_{\pm0.14}$ & \underline{2.21}$_{\pm0.08}$ & 2.16$_{\pm0.05}$ & \textbf{2.58}$_{\pm0.1}$ \\
S-C100/CNet & 0$_{\pm0}$ & 1.57$_{\pm0.05}$ & 0.54$_{\pm0.14}$ & 1.25$_{\pm0.12}$ & 1.67$_{\pm0.12}$ & 2.48$_{\pm0.12}$ & \underline{2.49}$_{\pm0.1}$ & \textbf{3.49}$_{\pm0.06}$ \\
S-SVHN/CNet & 0$_{\pm0}$ & 0.86$_{\pm0.05}$ & 0.49$_{\pm0.08}$ & 0.88$_{\pm0.06}$ & 1.13$_{\pm0.03}$ & 1.39$_{\pm0.04}$ & \underline{1.58}$_{\pm0.06}$ & \textbf{1.75}$_{\pm0.03}$ \\
S-TinyI/CNet & 0$_{\pm0}$ & 0.27$_{\pm0.05}$ & 0.08$_{\pm0.03}$ & 0.14$_{\pm0.02}$ & 0.22$_{\pm0.03}$ & 0.29$_{\pm0.03}$ & \underline{0.47}$_{\pm0.01}$ & \textbf{0.66}$_{\pm0.04}$ \\
CLEAR10/RNet & 0$_{\pm0}$ & 0.38$_{\pm0.13}$ & 0.46$_{\pm0.08}$ & 1.04$_{\pm0.06}$ & 1.4$_{\pm0.04}$ & 1.8$_{\pm0.06}$ & \underline{1.92}$_{\pm0.05}$ & \textbf{2.12}$_{\pm0.05}$ \\
CLEAR10/MoNet & 0$_{\pm0}$ & 1.03$_{\pm0.64}$ & 0.65$_{\pm0.23}$ & 2.31$_{\pm0.15}$ & 2.65$_{\pm0.49}$ & \underline{4.25}$_{\pm0.09}$ & 3.82$_{\pm0.26}$ & \textbf{5.34}$_{\pm0.11}$ \\
CLEAR100/RNet & 0$_{\pm0}$ & 2.76$_{\pm0.1}$ & 1.36$_{\pm0.22}$ & 2.52$_{\pm0.21}$ & 3.3$_{\pm0.23}$ & 3.85$_{\pm0.2}$ & \underline{3.98}$_{\pm0.19}$ & \textbf{4.24}$_{\pm0.22}$ \\
CLEAR100/MoNet & 0$_{\pm0}$ & 3.11$_{\pm0.53}$ & 1.26$_{\pm0.12}$ & 3.03$_{\pm0.52}$ & 4.24$_{\pm0.12}$ & 5.66$_{\pm0.19}$ & \underline{5.88}$_{\pm0.58}$ & \textbf{7.42}$_{\pm0.69}$ \\
Covertype/MLP & 0$_{\pm0}$ & 0.62$_{\pm0.14}$ & 0.24$_{\pm0.12}$ & 0.6$_{\pm0.16}$ & 0.83$_{\pm0.16}$ & \textbf{0.92}$_{\pm0.16}$ & 0.82$_{\pm0.13}$ & \underline{0.89}$_{\pm0.08}$ \\
    \bottomrule
    \end{tabular}%
    }
    \label{tab:pipeline-ablation}%
  \end{minipage}%
  \hfill\vline\hfill
  \begin{minipage}{.463\linewidth}
    \centering
  \caption{\\
  Online Accuracy differences between Ferret with and without \colorbox{orange!20}{gradients compensation algorithms}.
  \\ \\}
  \resizebox{\linewidth}{!}{%
    \begin{tabular}{|rrrr|rrrrl}
    \toprule
    \multicolumn{4}{c}{\cellcolor{orange!20}Ferret$_{\mathrm{M+}}$} & \multicolumn{4}{c}{\cellcolor{orange!20}Ferret$_{\mathrm{M}}$} \\
    \midrule
    \cellcolor{orange!20}Step-Aware & \cellcolor{orange!20}Gap-Aware & \cellcolor{orange!20}Fisher & \cellcolor{orange!20}Iter-Fisher & \cellcolor{orange!20}Step-Aware & \cellcolor{orange!20}Gap-Aware & \cellcolor{orange!20}Fisher & \cellcolor{orange!20}Iter-Fisher \\
    \midrule
    -56.04$_\pm{2.78}$ & -14.03$_\pm{1.24}$ & \underline{-0.02}$_\pm{0.01}$ & \textbf{0.01}$_\pm{0}$ & -43.53$_\pm{2.36}$ & -12.11$_\pm{1.03}$ & \underline{-0.01}$_\pm{0.01}$ & \textbf{0.02}$_\pm{0}$ \\
    -37.75$_\pm{2.17}$ & -9.07$_\pm{0.63}$ & \underline{-0.02}$_\pm{0.03}$ & \textbf{0.05}$_\pm{0.01}$ & -37.74$_\pm{2.53}$ & -7.07$_\pm{0.55}$ & \underline{-0.01}$_\pm{0.01}$ & \textbf{0.02}$_\pm{0}$ \\
    -20.5$_\pm{0.05}$ & -4.81$_\pm{0.15}$ & \underline{0.01}$_\pm{0.02}$ & \textbf{0.04}$_\pm{0.02}$ & -33.36$_\pm{0.13}$ & -3.46$_\pm{0.33}$ & \underline{0.01}$_\pm{0.01}$ & \textbf{0.04}$_\pm{0.02}$ \\
    -10.12$_\pm{0.19}$ & -1.71$_\pm{0.43}$ & \underline{-0.32}$_\pm{0.07}$ & \textbf{0.25}$_\pm{0.06}$ & -9.6$_\pm{0.19}$ & -1.22$_\pm{0.39}$ & \underline{-0.14}$_\pm{0.3}$ & \textbf{0.42}$_\pm{0.21}$ \\
    -8.08$_\pm{0.17}$ & -2.17$_\pm{0.05}$ & \underline{-0.04}$_\pm{0.08}$ & \textbf{0.13}$_\pm{0.05}$ & -5.4$_\pm{0.07}$ & -1.04$_\pm{0.04}$ & \underline{-0.01}$_\pm{0.07}$ & \textbf{0.1}$_\pm{0.02}$ \\
    -14.92$_\pm{0.42}$ & -2.63$_\pm{0.04}$ & \underline{0.02}$_\pm{0.04}$ & \textbf{0.31}$_\pm{0.2}$ & -24.7$_\pm{1.18}$ & -2.91$_\pm{0.11}$ & \underline{0.22}$_\pm{0.07}$ & \textbf{0.3}$_\pm{0.07}$ \\
    -3.72$_\pm{0.16}$ & -1.06$_\pm{0.09}$ & \underline{-0.01}$_\pm{0.02}$ & \textbf{0.06}$_\pm{0.03}$ & -1.32$_\pm{0.17}$ & -0.11$_\pm{0.19}$ & \underline{0.17}$_\pm{0.19}$ & \textbf{0.35}$_\pm{0.15}$ \\
    -23.93$_\pm{0.16}$ & -3.27$_\pm{0.05}$ & \underline{-0.22}$_\pm{0.11}$ & \textbf{0.1}$_\pm{0.08}$ & -33.41$_\pm{0.24}$ & -4.18$_\pm{0.34}$ & \underline{0.02}$_\pm{0.12}$ & \textbf{0.34}$_\pm{0.07}$ \\
    -24.56$_\pm{0.22}$ & -3.91$_\pm{0.2}$ & \underline{0.22}$_\pm{0.08}$ & \textbf{0.32}$_\pm{0.06}$ & -23.77$_\pm{0.3}$ & -3.15$_\pm{0.53}$ & \underline{0.23}$_\pm{0.12}$ & \textbf{0.39}$_\pm{0.1}$ \\
    -26.8$_\pm{3.2}$ & -4.21$_\pm{0.21}$ & \underline{-0.05}$_\pm{0.02}$ & \textbf{0.03}$_\pm{0.01}$ & -46.24$_\pm{1.41}$ & -4.82$_\pm{0.48}$ & \underline{-0.03}$_\pm{0.01}$ & \textbf{0.02}$_\pm{0}$ \\
    -14.43$_\pm{2.83}$ & -2.37$_\pm{0.14}$ & \underline{0.01}$_\pm{0.02}$ & \textbf{0.03}$_\pm{0.02}$ & -46.07$_\pm{4.03}$ & -2.1$_\pm{0.33}$ & \underline{-0.01}$_\pm{0.01}$ & \textbf{0}$_\pm{0}$ \\
    -6.93$_\pm{0.12}$ & -1$_\pm{0.14}$ & \underline{-0.19}$_\pm{0.09}$ & \textbf{0.1}$_\pm{0.03}$ & -8.11$_\pm{0.75}$ & -0.99$_\pm{0.24}$ & \underline{-0.12}$_\pm{0.12}$ & \textbf{0.23}$_\pm{0.12}$ \\
    -14.14$_\pm{0.37}$ & -3.05$_\pm{0.1}$ & \underline{-0.23}$_\pm{0.18}$ & \textbf{0.38}$_\pm{0.23}$ & -12.56$_\pm{0.23}$ & -1.92$_\pm{0.09}$ & \underline{-0.1}$_\pm{0.19}$ & \textbf{0.24}$_\pm{0.09}$ \\
    -5.69$_\pm{0.46}$ & -1.18$_\pm{0.12}$ & \underline{-0.03}$_\pm{0.03}$ & \textbf{0.05}$_\pm{0.03}$ & -12.13$_\pm{0.87}$ & -1.27$_\pm{0.09}$ & \underline{-0.05}$_\pm{0.02}$ & \textbf{0.03}$_\pm{0.01}$ \\
    -3.61$_\pm{0.14}$ & -1.01$_\pm{0.05}$ & \underline{0.05}$_\pm{0.1}$ & \textbf{0.12}$_\pm{0.1}$ & -1.7$_\pm{0.12}$ & -0.37$_\pm{0.03}$ & \underline{0.08}$_\pm{0.07}$ & \textbf{0.18}$_\pm{0.08}$ \\
    -6.26$_\pm{0.18}$ & -0.72$_\pm{0.08}$ & \underline{0.02}$_\pm{0.06}$ & \textbf{0.14}$_\pm{0.04}$ & -11.25$_\pm{0.28}$ & -0.52$_\pm{0.05}$ & \underline{-0.02}$_\pm{0.04}$ & \textbf{0.08}$_\pm{0.04}$ \\
    -18.17$_\pm{0.26}$ & -1.7$_\pm{0.05}$ & \underline{-0.14}$_\pm{0.32}$ & \textbf{0.5}$_\pm{0.15}$ & -20.69$_\pm{0.58}$ & -1.88$_\pm{0.46}$ & \underline{-0.28}$_\pm{0.37}$ & \textbf{0.36}$_\pm{0.18}$ \\
    -17.45$_\pm{0.02}$ & -0.38$_\pm{0.54}$ & \underline{-0.07}$_\pm{0.04}$ & \textbf{0.65}$_\pm{0.32}$ & -24.62$_\pm{0.24}$ & \underline{0.18}$_\pm{0.32}$ & -0.2$_\pm{0.18}$ & \textbf{1.27}$_\pm{0.21}$ \\
    -30.65$_\pm{0.93}$ & -0.3$_\pm{0.17}$ & \underline{0.63}$_\pm{0.46}$ & \textbf{1.2}$_\pm{0.48}$ & -26.85$_\pm{1.22}$ & -0.3$_\pm{0.33}$ & \underline{0.58}$_\pm{0.78}$ & \textbf{1.03}$_\pm{0.7}$ \\
    -9.2$_\pm{1.13}$ & -2.8$_\pm{0.1}$ & \underline{-0.09}$_\pm{0.05}$ & \textbf{0.05}$_\pm{0.02}$ & -10.27$_\pm{0.39}$ & -1.52$_\pm{0.14}$ & \underline{-0.01}$_\pm{0}$ & \textbf{0.09}$_\pm{0.01}$ \\
    \bottomrule
    \end{tabular}%
    }
  \label{tab:gc-ablation}%
  \end{minipage}%
\end{table*}%

\subsection{Overall Comparisons}
\label{sec:overall-comparisons}

Table~\ref{tab:overall-table} shows $agm_{\mathcal{B}}(\mathcal{A}, T)$ across 20 different settings to evaluate both performance and consumed memory of different frameworks.
Here, $\mathcal{B}$ is chosen to be the 1-Skip due to its low memory footprint.
From the table, it is evident that Ferret$_{\mathrm{M}}$ and Ferret$_{\mathrm{M+}}$ constantly outperform other competing algorithms.
Notably, Ferret$_{\mathrm{M+}}$ even achieves comparable performance compared to Oracle, indicating that Ferret effectively enables rapid adaptation.
On the other hand, while Ferret$_{\mathrm{M-}}$ shows slightly inferior performance compared to its counterparts, it demands less memory for OCL, as depicted in Fig.~\ref{fig:overall-memory-comparison}.
This implies that in scenarios where memory is severely constrained, Ferret is the only method capable of learning.

Furthermore, various OCL algorithms are integrated on CORe50/ConvNet in Table~\ref{tab:underlying-A}.
It can be observed that Ferret not only mitigates catastrophic forgetting (\textit{i.e.}, increased $tagm$) but also markedly enhances online performance (\textit{i.e.}, increased $agm$), validating its orthogonality and superiority compared to other OCL frameworks for rapid adaptation.

\subsection{Comparisons on Pipeline Parallelism}
\label{sec:pp-comparisons}

Table~\ref{tab:pipeline-ablation} compares $agm_{\mathcal{B}}(\mathcal{A}, T)$ of different pipeline parallelism strategies across 20 different settings to evaluate the performance of Ferret's fine-grained pipeline parallelism under memory constraints.
Specifically, $\mathcal{B}$ is selected as DAPPLE, and no gradients compensation is applied to any asynchronous pipeline parallelism strategies.
Additionally, Hanayo$_{\mathrm{1W}}$, Hanayo$_{\mathrm{2W}}$ and Hanayo$_{\mathrm{3W}}$ are three variants with 1, 2, and 3 waves, respectively.

In general, all asynchronous pipeline parallelism strategies significantly outperform synchronous pipeline parallelism strategies, even ZB, which claims to eliminate pipeline bubbles.
This is because synchronous pipeline parallelism strategies, in an effort to achieve higher hardware utilization rates and avoid conflicting model versions, must design complex workflows that stage gradients and update model parameters synchronously, resulting in delays in data processing and wasted data value. 
Conversely, asynchronous pipeline parallelism strategies process data and update model parameters immediately, thereby minimizing processing latency.
Among all asynchronous pipeline parallelism strategies, Ferret$_{\mathrm{M}}$'s fine-grained pipeline parallelism strategy consistently surpasses the others due to its more efficient memory utilization.

To investigate the impact of different pipeline configurations for Ferret, we select five different memory constraints ranging from minimum to maximum to simulate learning under varying memory budgets
Fig.~\ref{fig:pp-memory-comparison} shows that Ferret successfully solves Eq.~\ref{eq:main-obj} for obtaining optimal pipeline configurations under dynamic environments, scaling effectively as we increase the memory constraint.
Specifically, lack of precise control over each pipeline stage to balance between performance and memory footprint prevents competing strategies from scaling well.

\subsection{Comparisons on Gradients Compensation}
\label{sec:gc-comparisons}

To evaluate the effectiveness of Iter-Fisher, we apply various gradients compensation algorithms to Ferret$_{\mathrm{M+}}$ and Ferret$_{\mathrm{M}}$, and compare the final online accuracy gain.
The results are shown in Table~\ref{tab:gc-ablation}.
From the table, we can observe that applying Step-Aware and Gap-Aware algorithms for compensating stale gradients significantly reduces the online accuracy.
This is because these algorithms mitigate the gradient staleness problem by simply penalizing the step size of stale gradients, leading to a slow convergence rate when the system is highly paralleled.
Although Fisher leverages first-order information for better compensation, it does not consider varying levels of staleness at different stages of pipeline parallelism, resulting in a marginal decrease in accuracy compared to no compensation.
On the other hand, Iter-Fisher consistently improves online accuracy across all settings, without requiring manual hyper-parameters tuning.
This indicates that Iter-Fisher effectively adapts to different levels of staleness in parallelism, and automatically optimizes $\lambda$ for better compensation, demonstrating its robustness and effectiveness.

\section{Conclusion}
\label{sec:conclude}

This paper introduces Ferret, a novel framework designed to boost online accuracy of OCL algorithms under varying memory constraints. 
Ferret employs a fine-grained pipeline parallelism strategy to adapt to varying distributions of incoming streaming data rapidly.
To mitigate the gradient staleness problem in parallel processing, Ferret integrates an iterative gradient compensation algorithm to prevent performance degradation.
Additionally, pipelines are automatically scheduled to improve performance under any memory scenario by optimizing a bi-level optimization problem.
Extensive experiments conducted on 18 datasets and 5 models confirm Ferret’s superior efficiency and robustness compared to existing methods, demonstrating its potential as a scalable solution for adaptive, memory-efficient OCL.

\section*{Acknowledgments}
This work was supported in part by the National Natural Science Foundation of China under Grant 62306198; in part by National Major Scientific Instruments and Equipments Development Project of National Natural Science Foundation of China under Grant 62427820; and in part by the Natural Science Foundation of Sichuan Province under Grant 2024NSFSC1468.

{
    \small
    \bibliographystyle{ieeenat_fullname}
    \bibliography{main}

\begin{thebibliography}{90}
\providecommand{\natexlab}[1]{#1}
\providecommand{\url}[1]{\texttt{#1}}
\expandafter\ifx\csname urlstyle\endcsname\relax
  \providecommand{\doi}[1]{doi: #1}\else
  \providecommand{\doi}{doi: \begingroup \urlstyle{rm}\Url}\fi

\bibitem[Abadi et~al.(2016)Abadi, Barham, Chen, Chen, Davis, Dean, Devin, Ghemawat, Irving, Isard, et~al.]{abadi2016tensorflow}
Mart{\'\i}n Abadi, Paul Barham, Jianmin Chen, Zhifeng Chen, Andy Davis, Jeffrey Dean, Matthieu Devin, Sanjay Ghemawat, Geoffrey Irving, Michael Isard, et~al.
\newblock $\{$TensorFlow$\}$: a system for $\{$Large-Scale$\}$ machine learning.
\newblock In \emph{12th USENIX symposium on operating systems design and implementation (OSDI 16)}, pages 265--283, 2016.

\bibitem[Aljundi et~al.(2018)Aljundi, Babiloni, Elhoseiny, Rohrbach, and Tuytelaars]{aljundi2018memory}
Rahaf Aljundi, Francesca Babiloni, Mohamed Elhoseiny, Marcus Rohrbach, and Tinne Tuytelaars.
\newblock Memory aware synapses: Learning what (not) to forget.
\newblock In \emph{Proceedings of the European conference on computer vision (ECCV)}, pages 139--154, 2018.

\bibitem[Aljundi et~al.(2019{\natexlab{a}})Aljundi, Belilovsky, Tuytelaars, Charlin, Caccia, Lin, and Page-Caccia]{aljundi2019online}
Rahaf Aljundi, Eugene Belilovsky, Tinne Tuytelaars, Laurent Charlin, Massimo Caccia, Min Lin, and Lucas Page-Caccia.
\newblock Online continual learning with maximal interfered retrieval.
\newblock \emph{Advances in neural information processing systems}, 32, 2019{\natexlab{a}}.

\bibitem[Aljundi et~al.(2019{\natexlab{b}})Aljundi, Lin, Goujaud, and Bengio]{aljundi2019gradient}
Rahaf Aljundi, Min Lin, Baptiste Goujaud, and Yoshua Bengio.
\newblock Gradient based sample selection for online continual learning.
\newblock \emph{Advances in neural information processing systems}, 32, 2019{\natexlab{b}}.

\bibitem[Ansel et~al.(2024)Ansel, Yang, He, Gimelshein, Jain, Voznesensky, Bao, Bell, Berard, Burovski, et~al.]{ansel2024pytorch}
Jason Ansel, Edward Yang, Horace He, Natalia Gimelshein, Animesh Jain, Michael Voznesensky, Bin Bao, Peter Bell, David Berard, Evgeni Burovski, et~al.
\newblock Pytorch 2: Faster machine learning through dynamic python bytecode transformation and graph compilation.
\newblock 2024.

\bibitem[Barber et~al.(2021)Barber, Yu, Zamore, Lin, Jazayeri, Erlich, Savor, and Stumm]{barber2021bladerunner}
Jeff Barber, Ximing Yu, Laney~Kuenzel Zamore, Jerry Lin, Vahid Jazayeri, Shie Erlich, Tony Savor, and Michael Stumm.
\newblock Bladerunner: Stream processing at scale for a live view of backend data mutations at the edge.
\newblock In \emph{Proceedings of the ACM SIGOPS 28th Symposium on Operating Systems Principles}, pages 708--723, 2021.

\bibitem[Barkai et~al.(2019)Barkai, Hakimi, and Schuster]{barkai2019gap}
Saar Barkai, Ido Hakimi, and Assaf Schuster.
\newblock Gap aware mitigation of gradient staleness.
\newblock \emph{arXiv preprint arXiv:1909.10802}, 2019.

\bibitem[Blackard(1998)]{misc_covertype_31}
Jock Blackard.
\newblock {Covertype}.
\newblock UCI Machine Learning Repository, 1998.
\newblock {DOI}: https://doi.org/10.24432/C50K5N.

\bibitem[Buzzega et~al.(2020)Buzzega, Boschini, Porrello, Abati, and Calderara]{buzzega2020dark}
Pietro Buzzega, Matteo Boschini, Angelo Porrello, Davide Abati, and Simone Calderara.
\newblock Dark experience for general continual learning: a strong, simple baseline.
\newblock \emph{Advances in neural information processing systems}, 33:\penalty0 15920--15930, 2020.

\bibitem[Caccia et~al.(2021)Caccia, Aljundi, Asadi, Tuytelaars, Pineau, and Belilovsky]{caccia2021new}
Lucas Caccia, Rahaf Aljundi, Nader Asadi, Tinne Tuytelaars, Joelle Pineau, and Eugene Belilovsky.
\newblock New insights on reducing abrupt representation change in online continual learning.
\newblock \emph{arXiv preprint arXiv:2104.05025}, 2021.

\bibitem[Cai et~al.(2021)Cai, Sener, and Koltun]{cai2021online}
Zhipeng Cai, Ozan Sener, and Vladlen Koltun.
\newblock Online continual learning with natural distribution shifts: An empirical study with visual data.
\newblock In \emph{Proceedings of the IEEE/CVF international conference on computer vision}, pages 8281--8290, 2021.

\bibitem[Chaudhry et~al.(2019)Chaudhry, Rohrbach, Elhoseiny, Ajanthan, Dokania, Torr, and Ranzato]{chaudhry2019continual}
Arslan Chaudhry, Marcus Rohrbach, Mohamed Elhoseiny, Thalaiyasingam Ajanthan, P Dokania, P Torr, and M Ranzato.
\newblock Continual learning with tiny episodic memories.
\newblock In \emph{Workshop on Multi-Task and Lifelong Reinforcement Learning}, 2019.

\bibitem[Chen et~al.(2016)Chen, Xu, Zhang, and Guestrin]{chen2016training}
Tianqi Chen, Bing Xu, Chiyuan Zhang, and Carlos Guestrin.
\newblock Training deep nets with sublinear memory cost.
\newblock \emph{arXiv preprint arXiv:1604.06174}, 2016.

\bibitem[Chen et~al.(2022)Chen, Xie, Ma, Gu, Peng, Lin, Wu, and Zhu]{chen2022sapipe}
Yangrui Chen, Cong Xie, Meng Ma, Juncheng Gu, Yanghua Peng, Haibin Lin, Chuan Wu, and Yibo Zhu.
\newblock Sapipe: Staleness-aware pipeline for data parallel dnn training.
\newblock \emph{Advances in Neural Information Processing Systems}, 35:\penalty0 17981--17993, 2022.

\bibitem[Chollet(2015)]{chollet2015}
François Chollet.
\newblock keras.
\newblock \url{https://github.com/fchollet/keras}, 2015.

\bibitem[Cohen et~al.(2017)Cohen, Afshar, Tapson, and Van~Schaik]{cohen2017emnist}
Gregory Cohen, Saeed Afshar, Jonathan Tapson, and Andre Van~Schaik.
\newblock Emnist: Extending mnist to handwritten letters.
\newblock In \emph{2017 international joint conference on neural networks (IJCNN)}, pages 2921--2926. IEEE, 2017.

\bibitem[Darvish~Rouhani et~al.(2019)Darvish~Rouhani, Chen, and Koushanfar]{darvish2019deepsigns}
Bita Darvish~Rouhani, Huili Chen, and Farinaz Koushanfar.
\newblock Deepsigns: An end-to-end watermarking framework for ownership protection of deep neural networks.
\newblock In \emph{Proceedings of the twenty-fourth international conference on architectural support for programming languages and operating systems}, pages 485--497, 2019.

\bibitem[de~Assuncao et~al.(2018)de~Assuncao, da~Silva~Veith, and Buyya]{de2018distributed}
Marcos~Dias de Assuncao, Alexandre da Silva~Veith, and Rajkumar Buyya.
\newblock Distributed data stream processing and edge computing: A survey on resource elasticity and future directions.
\newblock \emph{Journal of Network and Computer Applications}, 103:\penalty0 1--17, 2018.

\bibitem[De~Lange and Tuytelaars(2021)]{de2021continual2}
Matthias De~Lange and Tinne Tuytelaars.
\newblock Continual prototype evolution: Learning online from non-stationary data streams.
\newblock In \emph{Proceedings of the IEEE/CVF international conference on computer vision}, pages 8250--8259, 2021.

\bibitem[De~Lange et~al.(2021)De~Lange, Aljundi, Masana, Parisot, Jia, Leonardis, Slabaugh, and Tuytelaars]{de2021continual}
Matthias De~Lange, Rahaf Aljundi, Marc Masana, Sarah Parisot, Xu Jia, Ale{\v{s}} Leonardis, Gregory Slabaugh, and Tinne Tuytelaars.
\newblock A continual learning survey: Defying forgetting in classification tasks.
\newblock \emph{IEEE transactions on pattern analysis and machine intelligence}, 44\penalty0 (7):\penalty0 3366--3385, 2021.

\bibitem[Deng et~al.(2009)Deng, Dong, Socher, Li, Li, and Fei-Fei]{deng2009imagenet}
Jia Deng, Wei Dong, Richard Socher, Li-Jia Li, Kai Li, and Li Fei-Fei.
\newblock Imagenet: A large-scale hierarchical image database.
\newblock In \emph{2009 IEEE conference on computer vision and pattern recognition}, pages 248--255. Ieee, 2009.

\bibitem[Deng(2012)]{deng2012mnist}
Li Deng.
\newblock The mnist database of handwritten digit images for machine learning research.
\newblock \emph{IEEE Signal Processing Magazine}, 29\penalty0 (6):\penalty0 141--142, 2012.

\bibitem[Dixon et~al.(2020)Dixon, Halperin, and Bilokon]{dixon2020machine}
Matthew~F Dixon, Igor Halperin, and Paul Bilokon.
\newblock \emph{Machine learning in finance}.
\newblock Springer, 2020.

\bibitem[Fan et~al.(2021)Fan, Rong, Meng, Cao, Wang, Zheng, Wu, Long, Yang, Xia, et~al.]{fan2021dapple}
Shiqing Fan, Yi Rong, Chen Meng, Zongyan Cao, Siyu Wang, Zhen Zheng, Chuan Wu, Guoping Long, Jun Yang, Lixue Xia, et~al.
\newblock Dapple: A pipelined data parallel approach for training large models.
\newblock In \emph{Proceedings of the 26th ACM SIGPLAN Symposium on Principles and Practice of Parallel Programming}, pages 431--445, 2021.

\bibitem[Fernando et~al.(2017)Fernando, Banarse, Blundell, Zwols, Ha, Rusu, Pritzel, and Wierstra]{fernando2017pathnet}
Chrisantha Fernando, Dylan Banarse, Charles Blundell, Yori Zwols, David Ha, Andrei~A Rusu, Alexander Pritzel, and Daan Wierstra.
\newblock Pathnet: Evolution channels gradient descent in super neural networks.
\newblock \emph{arXiv preprint arXiv:1701.08734}, 2017.

\bibitem[Fini et~al.(2020)Fini, Lathuiliere, Sangineto, Nabi, and Ricci]{fini2020online}
Enrico Fini, St{\'e}phane Lathuiliere, Enver Sangineto, Moin Nabi, and Elisa Ricci.
\newblock Online continual learning under extreme memory constraints.
\newblock In \emph{Computer Vision--ECCV 2020: 16th European Conference, Glasgow, UK, August 23--28, 2020, Proceedings, Part XXVIII 16}, pages 720--735. Springer, 2020.

\bibitem[Finn et~al.(2019)Finn, Rajeswaran, Kakade, and Levine]{finn2019online}
Chelsea Finn, Aravind Rajeswaran, Sham Kakade, and Sergey Levine.
\newblock Online meta-learning.
\newblock In \emph{International conference on machine learning}, pages 1920--1930. PMLR, 2019.

\bibitem[Friedman et~al.(2001)Friedman, Hastie, and Tibshirani]{friedman2001elements}
Jerome Friedman, Trevor Hastie, and Robert Tibshirani.
\newblock \emph{The elements of statistical learning}.
\newblock Springer series in statistics New York, 2001.

\bibitem[Ghunaim et~al.(2023)Ghunaim, Bibi, Alhamoud, Alfarra, Al~Kader~Hammoud, Prabhu, Torr, and Ghanem]{ghunaim2023real}
Yasir Ghunaim, Adel Bibi, Kumail Alhamoud, Motasem Alfarra, Hasan~Abed Al~Kader~Hammoud, Ameya Prabhu, Philip~HS Torr, and Bernard Ghanem.
\newblock Real-time evaluation in online continual learning: A new hope.
\newblock In \emph{Proceedings of the IEEE/CVF Conference on Computer Vision and Pattern Recognition}, pages 11888--11897, 2023.

\bibitem[Girdhar et~al.(2023)Girdhar, El-Nouby, Liu, Singh, Alwala, Joulin, and Misra]{girdhar2023imagebind}
Rohit Girdhar, Alaaeldin El-Nouby, Zhuang Liu, Mannat Singh, Kalyan~Vasudev Alwala, Armand Joulin, and Ishan Misra.
\newblock Imagebind: One embedding space to bind them all.
\newblock In \emph{Proceedings of the IEEE/CVF Conference on Computer Vision and Pattern Recognition}, pages 15180--15190, 2023.

\bibitem[Guan et~al.(2023)Guan, Qiu, Leng, Yang, Yu, Liu, Feng, Zhu, Zhou, Liang, et~al.]{guan2023amanda}
Yue Guan, Yuxian Qiu, Jingwen Leng, Fan Yang, Shuo Yu, Yunxin Liu, Yu Feng, Yuhao Zhu, Lidong Zhou, Yun Liang, et~al.
\newblock Amanda: Unified instrumentation framework for deep neural networks.
\newblock 2023.

\bibitem[Gunasekara et~al.(2023)Gunasekara, Pfahringer, Gomes, and Bifet]{gunasekara2023survey}
Nuwan Gunasekara, Bernhard Pfahringer, Heitor~Murilo Gomes, and Albert Bifet.
\newblock Survey on online streaming continual learning.
\newblock In \emph{Proceedings of the Thirty-Second International Joint Conference on Artificial Intelligence, IJCAI 2023, 19th-25th August 2023, Macao, SAR, China}, pages 6628--6637. ijcai. org, 2023.

\bibitem[Hardy et~al.(2017)Hardy, Le~Merrer, and Sericola]{hardy2017distributed}
Corentin Hardy, Erwan Le~Merrer, and Bruno Sericola.
\newblock Distributed deep learning on edge-devices: feasibility via adaptive compression.
\newblock In \emph{2017 IEEE 16th international symposium on network computing and applications (NCA)}, pages 1--8. IEEE, 2017.

\bibitem[He et~al.(2016)He, Zhang, Ren, and Sun]{he2016deep}
Kaiming He, Xiangyu Zhang, Shaoqing Ren, and Jian Sun.
\newblock Deep residual learning for image recognition.
\newblock In \emph{Proceedings of the IEEE conference on computer vision and pattern recognition}, pages 770--778, 2016.

\bibitem[Howard et~al.(2017)Howard, Zhu, Chen, Kalenichenko, Wang, Weyand, Andreetto, and Adam]{howard2017mobilenets}
Andrew~G Howard, Menglong Zhu, Bo Chen, Dmitry Kalenichenko, Weijun Wang, Tobias Weyand, Marco Andreetto, and Hartwig Adam.
\newblock Mobilenets: Efficient convolutional neural networks for mobile vision applications.
\newblock \emph{arXiv preprint arXiv:1704.04861}, 2017.

\bibitem[Hu et~al.(2023)Hu, Ye, Zhang, Chen, Sun, Wen, and Zhang]{hu2023hydro}
Qinghao Hu, Zhisheng Ye, Meng Zhang, Qiaoling Chen, Peng Sun, Yonggang Wen, and Tianwei Zhang.
\newblock Hydro:$\{$Surrogate-Based$\}$ hyperparameter tuning service in datacenters.
\newblock In \emph{17th USENIX Symposium on Operating Systems Design and Implementation (OSDI 23)}, pages 757--777, 2023.

\bibitem[Hu et~al.(2020)Hu, Liang, Li, Deng, Zuo, Ji, Xie, Ding, Liu, Sherwood, et~al.]{hu2020deepsniffer}
Xing Hu, Ling Liang, Shuangchen Li, Lei Deng, Pengfei Zuo, Yu Ji, Xinfeng Xie, Yufei Ding, Chang Liu, Timothy Sherwood, et~al.
\newblock Deepsniffer: A dnn model extraction framework based on learning architectural hints.
\newblock In \emph{Proceedings of the Twenty-Fifth International Conference on Architectural Support for Programming Languages and Operating Systems}, pages 385--399, 2020.

\bibitem[Huang et~al.(2019)Huang, Cheng, Bapna, Firat, Chen, Chen, Lee, Ngiam, Le, Wu, et~al.]{huang2019gpipe}
Yanping Huang, Youlong Cheng, Ankur Bapna, Orhan Firat, Dehao Chen, Mia Chen, HyoukJoong Lee, Jiquan Ngiam, Quoc~V Le, Yonghui Wu, et~al.
\newblock Gpipe: Efficient training of giant neural networks using pipeline parallelism.
\newblock \emph{Advances in neural information processing systems}, 32, 2019.

\bibitem[Isele and Cosgun(2018)]{isele2018selective}
David Isele and Akansel Cosgun.
\newblock Selective experience replay for lifelong learning.
\newblock In \emph{Proceedings of the AAAI Conference on Artificial Intelligence}, 2018.

\bibitem[Jayarajan et~al.(2021)Jayarajan, Hau, Goodwin, and Pekhimenko]{jayarajan2021lifestream}
Anand Jayarajan, Kimberly Hau, Andrew Goodwin, and Gennady Pekhimenko.
\newblock Lifestream: a high-performance stream processing engine for periodic streams.
\newblock In \emph{Proceedings of the 26th ACM International Conference on Architectural Support for Programming Languages and Operating Systems}, pages 107--122, 2021.

\bibitem[Jiang et~al.(2017)Jiang, Cui, Zhang, and Yu]{jiang2017heterogeneity}
Jiawei Jiang, Bin Cui, Ce Zhang, and Lele Yu.
\newblock Heterogeneity-aware distributed parameter servers.
\newblock In \emph{Proceedings of the 2017 ACM International Conference on Management of Data}, pages 463--478, 2017.

\bibitem[Kim et~al.(2016)Kim, Ho, Lee, Zheng, Dai, Gibson, and Xing]{kim2016strads}
Jin~Kyu Kim, Qirong Ho, Seunghak Lee, Xun Zheng, Wei Dai, Garth~A Gibson, and Eric~P Xing.
\newblock Strads: A distributed framework for scheduled model parallel machine learning.
\newblock In \emph{Proceedings of the Eleventh European Conference on Computer Systems}, pages 1--16, 2016.

\bibitem[Krizhevsky et~al.(2009)Krizhevsky, Hinton, et~al.]{krizhevsky2009learning}
Alex Krizhevsky, Geoffrey Hinton, et~al.
\newblock Learning multiple layers of features from tiny images.
\newblock 2009.

\bibitem[Kwon et~al.(2023)Kwon, Chauhan, Jia, Venieris, and Mascolo]{kwon2023lifelearner}
Young~D Kwon, Jagmohan Chauhan, Hong Jia, Stylianos~I Venieris, and Cecilia Mascolo.
\newblock Lifelearner: Hardware-aware meta continual learning system for embedded computing platforms.
\newblock \emph{arXiv preprint arXiv:2311.11420}, 2023.

\bibitem[Le and Yang(2015)]{le2015tiny}
Ya Le and Xuan Yang.
\newblock Tiny imagenet visual recognition challenge.
\newblock \emph{CS 231N}, 7\penalty0 (7):\penalty0 3, 2015.

\bibitem[Li et~al.(2022)Li, Shen, and Chen]{li2022camel}
Yiming Li, Yanyan Shen, and Lei Chen.
\newblock Camel: Managing data for efficient stream learning.
\newblock In \emph{Proceedings of the 2022 International Conference on Management of Data}, pages 1271--1285, 2022.

\bibitem[Li and Hoiem(2017)]{li2017learning}
Zhizhong Li and Derek Hoiem.
\newblock Learning without forgetting.
\newblock \emph{IEEE transactions on pattern analysis and machine intelligence}, 40\penalty0 (12):\penalty0 2935--2947, 2017.

\bibitem[Lin et~al.(2021)Lin, Shi, Pathak, and Ramanan]{lin2021clear}
Zhiqiu Lin, Jia Shi, Deepak Pathak, and Deva Ramanan.
\newblock The clear benchmark: Continual learning on real-world imagery.
\newblock In \emph{Thirty-fifth conference on neural information processing systems datasets and benchmarks track (round 2)}, 2021.

\bibitem[Liu et~al.(2023)Liu, Cheng, Zhou, and You]{liu2023hanayo}
Ziming Liu, Shenggan Cheng, Haotian Zhou, and Yang You.
\newblock Hanayo: Harnessing wave-like pipeline parallelism for enhanced large model training efficiency.
\newblock In \emph{Proceedings of the International Conference for High Performance Computing, Networking, Storage and Analysis}, pages 1--13, 2023.

\bibitem[Lomonaco and Maltoni(2017)]{lomonaco2017core50}
Vincenzo Lomonaco and Davide Maltoni.
\newblock Core50: a new dataset and benchmark for continuous object recognition.
\newblock In \emph{Conference on robot learning}, pages 17--26. PMLR, 2017.

\bibitem[Lopez-Paz and Ranzato(2017)]{lopez2017gradient}
David Lopez-Paz and Marc'Aurelio Ranzato.
\newblock Gradient episodic memory for continual learning.
\newblock \emph{Advances in neural information processing systems}, 30, 2017.

\bibitem[Mai et~al.(2022)Mai, Li, Jeong, Quispe, Kim, and Sanner]{mai2022online}
Zheda Mai, Ruiwen Li, Jihwan Jeong, David Quispe, Hyunwoo Kim, and Scott Sanner.
\newblock Online continual learning in image classification: An empirical survey.
\newblock \emph{Neurocomputing}, 469:\penalty0 28--51, 2022.

\bibitem[Menghani(2023)]{menghani2023efficient}
Gaurav Menghani.
\newblock Efficient deep learning: A survey on making deep learning models smaller, faster, and better.
\newblock \emph{ACM Computing Surveys}, 55\penalty0 (12):\penalty0 1--37, 2023.

\bibitem[Mirzasoleiman et~al.(2020)Mirzasoleiman, Bilmes, and Leskovec]{mirzasoleiman2020coresets}
Baharan Mirzasoleiman, Jeff Bilmes, and Jure Leskovec.
\newblock Coresets for data-efficient training of machine learning models.
\newblock In \emph{International Conference on Machine Learning}, pages 6950--6960. PMLR, 2020.

\bibitem[Moritz et~al.(2018)Moritz, Nishihara, Wang, Tumanov, Liaw, Liang, Elibol, Yang, Paul, Jordan, et~al.]{moritz2018ray}
Philipp Moritz, Robert Nishihara, Stephanie Wang, Alexey Tumanov, Richard Liaw, Eric Liang, Melih Elibol, Zongheng Yang, William Paul, Michael~I Jordan, et~al.
\newblock Ray: A distributed framework for emerging $\{$AI$\}$ applications.
\newblock In \emph{13th USENIX symposium on operating systems design and implementation (OSDI 18)}, pages 561--577, 2018.

\bibitem[Muhammad et~al.(2020)Muhammad, Ullah, Lloret, Del~Ser, and de~Albuquerque]{muhammad2020deep}
Khan Muhammad, Amin Ullah, Jaime Lloret, Javier Del~Ser, and Victor Hugo~C de Albuquerque.
\newblock Deep learning for safe autonomous driving: Current challenges and future directions.
\newblock \emph{IEEE Transactions on Intelligent Transportation Systems}, 22\penalty0 (7):\penalty0 4316--4336, 2020.

\bibitem[Murshed et~al.(2021)Murshed, Murphy, Hou, Khan, Ananthanarayanan, and Hussain]{murshed2021machine}
MG~Sarwar Murshed, Christopher Murphy, Daqing Hou, Nazar Khan, Ganesh Ananthanarayanan, and Faraz Hussain.
\newblock Machine learning at the network edge: A survey.
\newblock \emph{ACM Computing Surveys (CSUR)}, 54\penalty0 (8):\penalty0 1--37, 2021.

\bibitem[Narayanan et~al.(2019)Narayanan, Harlap, Phanishayee, Seshadri, Devanur, Ganger, Gibbons, and Zaharia]{narayanan2019pipedream}
Deepak Narayanan, Aaron Harlap, Amar Phanishayee, Vivek Seshadri, Nikhil~R Devanur, Gregory~R Ganger, Phillip~B Gibbons, and Matei Zaharia.
\newblock Pipedream: generalized pipeline parallelism for dnn training.
\newblock In \emph{Proceedings of the 27th ACM symposium on operating systems principles}, pages 1--15, 2019.

\bibitem[Narayanan et~al.(2021)Narayanan, Phanishayee, Shi, Chen, and Zaharia]{narayanan2021memory}
Deepak Narayanan, Amar Phanishayee, Kaiyu Shi, Xie Chen, and Matei Zaharia.
\newblock Memory-efficient pipeline-parallel dnn training.
\newblock In \emph{International Conference on Machine Learning}, pages 7937--7947. PMLR, 2021.

\bibitem[Netzer et~al.(2011)Netzer, Wang, Coates, Bissacco, Wu, Ng, et~al.]{netzer2011reading}
Yuval Netzer, Tao Wang, Adam Coates, Alessandro Bissacco, Baolin Wu, Andrew~Y Ng, et~al.
\newblock Reading digits in natural images with unsupervised feature learning.
\newblock In \emph{NIPS workshop on deep learning and unsupervised feature learning}, page~7. Granada, Spain, 2011.

\bibitem[Niu et~al.(2022)Niu, Wu, Zhang, Chen, Zheng, Zhao, and Tan]{niu2022efficient}
Shuaicheng Niu, Jiaxiang Wu, Yifan Zhang, Yaofo Chen, Shijian Zheng, Peilin Zhao, and Mingkui Tan.
\newblock Efficient test-time model adaptation without forgetting.
\newblock In \emph{International conference on machine learning}, pages 16888--16905. PMLR, 2022.

\bibitem[NVIDIA et~al.(2020)NVIDIA, Vingelmann, and Fitzek]{cuda}
NVIDIA, Péter Vingelmann, and Frank~H.P. Fitzek.
\newblock Cuda, release: 10.2.89, 2020.

\bibitem[Papadopoulos(2014)]{papadopoulos2014information}
Alecos Papadopoulos.
\newblock The information matrix equality: proof, misspecification, and the quasi-maximum likelihood estimator.
\newblock \emph{Athens University of Economics and Business}, 2014.

\bibitem[Paszke et~al.(2019)Paszke, Gross, Massa, Lerer, Bradbury, Chanan, Killeen, Lin, Gimelshein, Antiga, et~al.]{paszke2019pytorch}
Adam Paszke, Sam Gross, Francisco Massa, Adam Lerer, James Bradbury, Gregory Chanan, Trevor Killeen, Zeming Lin, Natalia Gimelshein, Luca Antiga, et~al.
\newblock Pytorch: An imperative style, high-performance deep learning library.
\newblock \emph{Advances in neural information processing systems}, 32, 2019.

\bibitem[Prabhu et~al.(2020)Prabhu, Torr, and Dokania]{prabhu2020gdumb}
Ameya Prabhu, Philip~HS Torr, and Puneet~K Dokania.
\newblock Gdumb: A simple approach that questions our progress in continual learning.
\newblock In \emph{Computer Vision--ECCV 2020: 16th European Conference, Glasgow, UK, August 23--28, 2020, Proceedings, Part II 16}, pages 524--540. Springer, 2020.

\bibitem[Qi et~al.(2023)Qi, Wan, Huang, and Lin]{qi2023zero}
Penghui Qi, Xinyi Wan, Guangxing Huang, and Min Lin.
\newblock Zero bubble pipeline parallelism.
\newblock \emph{arXiv preprint arXiv:2401.10241}, 2023.

\bibitem[Rasley et~al.(2020)Rasley, Rajbhandari, Ruwase, and He]{rasley2020deepspeed}
Jeff Rasley, Samyam Rajbhandari, Olatunji Ruwase, and Yuxiong He.
\newblock Deepspeed: System optimizations enable training deep learning models with over 100 billion parameters.
\newblock In \emph{Proceedings of the 26th ACM SIGKDD International Conference on Knowledge Discovery \& Data Mining}, pages 3505--3506, 2020.

\bibitem[Rebuffi et~al.(2017)Rebuffi, Kolesnikov, Sperl, and Lampert]{rebuffi2017icarl}
Sylvestre-Alvise Rebuffi, Alexander Kolesnikov, Georg Sperl, and Christoph~H Lampert.
\newblock icarl: Incremental classifier and representation learning.
\newblock In \emph{Proceedings of the IEEE conference on Computer Vision and Pattern Recognition}, pages 2001--2010, 2017.

\bibitem[Rusu et~al.(2016)Rusu, Rabinowitz, Desjardins, Soyer, Kirkpatrick, Kavukcuoglu, Pascanu, and Hadsell]{rusu2016progressive}
Andrei~A Rusu, Neil~C Rabinowitz, Guillaume Desjardins, Hubert Soyer, James Kirkpatrick, Koray Kavukcuoglu, Razvan Pascanu, and Raia Hadsell.
\newblock Progressive neural networks.
\newblock \emph{arXiv preprint arXiv:1606.04671}, 2016.

\bibitem[Sahoo et~al.(2017)Sahoo, Pham, Lu, and Hoi]{sahoo2017online}
Doyen Sahoo, Quang Pham, Jing Lu, and Steven~CH Hoi.
\newblock Online deep learning: Learning deep neural networks on the fly.
\newblock \emph{arXiv preprint arXiv:1711.03705}, 2017.

\bibitem[Sayce(2022)]{david2022tweets}
David Sayce.
\newblock The number of tweets per day in 2022.
\newblock 2022.

\bibitem[Sergeev and Del~Balso(2018)]{sergeev2018horovod}
Alexander Sergeev and Mike Del~Balso.
\newblock Horovod: fast and easy distributed deep learning in tensorflow.
\newblock \emph{arXiv preprint arXiv:1802.05799}, 2018.

\bibitem[Shoeybi et~al.(2019)Shoeybi, Patwary, Puri, LeGresley, Casper, and Catanzaro]{shoeybi2019megatron}
Mohammad Shoeybi, Mostofa Patwary, Raul Puri, Patrick LeGresley, Jared Casper, and Bryan Catanzaro.
\newblock Megatron-lm: Training multi-billion parameter language models using model parallelism.
\newblock \emph{arXiv preprint arXiv:1909.08053}, 2019.

\bibitem[Tai et~al.(2018)Tai, Sharan, Bailis, and Valiant]{tai2018sketching}
Kai~Sheng Tai, Vatsal Sharan, Peter Bailis, and Gregory Valiant.
\newblock Sketching linear classifiers over data streams.
\newblock In \emph{Proceedings of the 2018 international conference on management of data}, pages 757--772, 2018.

\bibitem[Touvron et~al.(2023)Touvron, Lavril, Izacard, Martinet, Lachaux, Lacroix, Rozi{\`e}re, Goyal, Hambro, Azhar, et~al.]{touvron2023llama}
Hugo Touvron, Thibaut Lavril, Gautier Izacard, Xavier Martinet, Marie-Anne Lachaux, Timoth{\'e}e Lacroix, Baptiste Rozi{\`e}re, Naman Goyal, Eric Hambro, Faisal Azhar, et~al.
\newblock Llama: Open and efficient foundation language models.
\newblock \emph{arXiv preprint arXiv:2302.13971}, 2023.

\bibitem[Valavi et~al.(2022)Valavi, Hestness, Ardalani, and Iansiti]{valavi2022time}
Ehsan Valavi, Joel Hestness, Newsha Ardalani, and Marco Iansiti.
\newblock Time and the value of data.
\newblock \emph{arXiv preprint arXiv:2203.09118}, 2022.

\bibitem[Veseli et~al.(2023)Veseli, Hammonds, Henke, Parraga, and Schwarz]{veseli2023streaming}
Sini{\v{s}}a Veseli, John Hammonds, Steven Henke, Hannah Parraga, and Nicholas Schwarz.
\newblock Streaming data from experimental facilities to supercomputers for real-time data processing.
\newblock In \emph{Proceedings of the SC'23 Workshops of The International Conference on High Performance Computing, Network, Storage, and Analysis}, pages 2110--2117, 2023.

\bibitem[Wang et~al.(2024)Wang, Zhang, Su, and Zhu]{wang2024comprehensive}
Liyuan Wang, Xingxing Zhang, Hang Su, and Jun Zhu.
\newblock A comprehensive survey of continual learning: Theory, method and application.
\newblock \emph{IEEE Transactions on Pattern Analysis and Machine Intelligence}, 2024.

\bibitem[Xiao et~al.(2017)Xiao, Rasul, and Vollgraf]{xiao2017fashion}
Han Xiao, Kashif Rasul, and Roland Vollgraf.
\newblock Fashion-mnist: a novel image dataset for benchmarking machine learning algorithms.
\newblock \emph{arXiv preprint arXiv:1708.07747}, 2017.

\bibitem[Xie et~al.(2020)Xie, Ren, Lu, Yang, Xu, Wu, Lin, Ao, Xu, and Shu]{xie2020kraken}
Minhui Xie, Kai Ren, Youyou Lu, Guangxu Yang, Qingxing Xu, Bihai Wu, Jiazhen Lin, Hongbo Ao, Wanhong Xu, and Jiwu Shu.
\newblock Kraken: memory-efficient continual learning for large-scale real-time recommendations.
\newblock In \emph{SC20: International Conference for High Performance Computing, Networking, Storage and Analysis}, pages 1--17. IEEE, 2020.

\bibitem[Yang et~al.(2023)Yang, Kang, and Mirzasoleiman]{yang2023towards}
Yu Yang, Hao Kang, and Baharan Mirzasoleiman.
\newblock Towards sustainable learning: Coresets for data-efficient deep learning.
\newblock In \emph{International Conference on Machine Learning}, pages 39314--39330. PMLR, 2023.

\bibitem[Yoon et~al.(2017)Yoon, Yang, Lee, and Hwang]{yoon2017lifelong}
Jaehong Yoon, Eunho Yang, Jeongtae Lee, and Sung~Ju Hwang.
\newblock Lifelong learning with dynamically expandable networks.
\newblock \emph{arXiv preprint arXiv:1708.01547}, 2017.

\bibitem[Yoon et~al.(2021)Yoon, Madaan, Yang, and Hwang]{yoon2021online}
Jaehong Yoon, Divyam Madaan, Eunho Yang, and Sung~Ju Hwang.
\newblock Online coreset selection for rehearsal-based continual learning.
\newblock \emph{arXiv preprint arXiv:2106.01085}, 2021.

\bibitem[Zhang et~al.(2020)Zhang, Han, Yang, Zhang, Liu, Yang, and Zhou]{zhang2020retiarii}
Quanlu Zhang, Zhenhua Han, Fan Yang, Yuge Zhang, Zhe Liu, Mao Yang, and Lidong Zhou.
\newblock Retiarii: A deep learning $\{$Exploratory-Training$\}$ framework.
\newblock In \emph{14th USENIX Symposium on Operating Systems Design and Implementation (OSDI 20)}, pages 919--936, 2020.

\bibitem[Zheng et~al.(2017)Zheng, Meng, Wang, Chen, Yu, Ma, and Liu]{zheng2017asynchronous}
Shuxin Zheng, Qi Meng, Taifeng Wang, Wei Chen, Nenghai Yu, Zhi-Ming Ma, and Tie-Yan Liu.
\newblock Asynchronous stochastic gradient descent with delay compensation.
\newblock In \emph{International conference on machine learning}, pages 4120--4129. PMLR, 2017.

\bibitem[Zheng et~al.(2019)Zheng, Oh, Zhai, Shen, Yi, and Chen]{zheng2019hiwaylib}
Zhen Zheng, Chanyoung Oh, Jidong Zhai, Xipeng Shen, Youngmin Yi, and Wenguang Chen.
\newblock Hiwaylib: A software framework for enabling high performance communications for heterogeneous pipeline computations.
\newblock In \emph{Proceedings of the Twenty-Fourth International Conference on Architectural Support for Programming Languages and Operating Systems}, pages 153--166, 2019.

\bibitem[Zhou et~al.(2021)Zhou, Ye, and Lv]{zhou2021communication}
Yuhao Zhou, Qing Ye, and Jiancheng Lv.
\newblock Communication-efficient federated learning with compensated overlap-fedavg.
\newblock \emph{IEEE Transactions on Parallel and Distributed Systems}, 33\penalty0 (1):\penalty0 192--205, 2021.

\bibitem[Zhou et~al.(2023)Zhou, Shi, Li, Sun, Ye, and Lv]{zhou2023communication}
Yuhao Zhou, Mingjia Shi, Yuanxi Li, Yanan Sun, Qing Ye, and Jiancheng Lv.
\newblock Communication-efficient federated learning with single-step synthetic features compressor for faster convergence.
\newblock In \emph{Proceedings of the IEEE/CVF International Conference on Computer Vision}, pages 5031--5040, 2023.

\bibitem[Zhou et~al.(2024)Zhou, Shi, Tian, Ye, and Lv]{zhou2024defta}
Yuhao Zhou, Minjia Shi, Yuxin Tian, Qing Ye, and Jiancheng Lv.
\newblock Defta: A plug-and-play peer-to-peer decentralized federated learning framework.
\newblock \emph{Information Sciences}, 670:\penalty0 120582, 2024.

\bibitem[Zhou et~al.(2025)Zhou, Tian, Shi, Li, Sun, Ye, and Lv]{zhou20253sfc}
Yuhao Zhou, Yuxin Tian, Mingjia Shi, Yuanxi Li, Yanan Sun, Qing Ye, and Jiancheng Lv.
\newblock E-3sfc: Communication-efficient federated learning with double-way features synthesizing.
\newblock \emph{IEEE Transactions on Neural Networks and Learning Systems}, \penalty0 (99):\penalty0 1--15, 2025.

\end{thebibliography}
}

\clearpage
\setcounter{page}{1}
\maketitlesupplementary

\section{Additional Related Work}
\label{sec:more-related}

\begin{figure}
    \centering
    \subfloat[Online continual learning\label{fig:vanilla-sl}]{\includegraphics[width=0.45\textwidth]{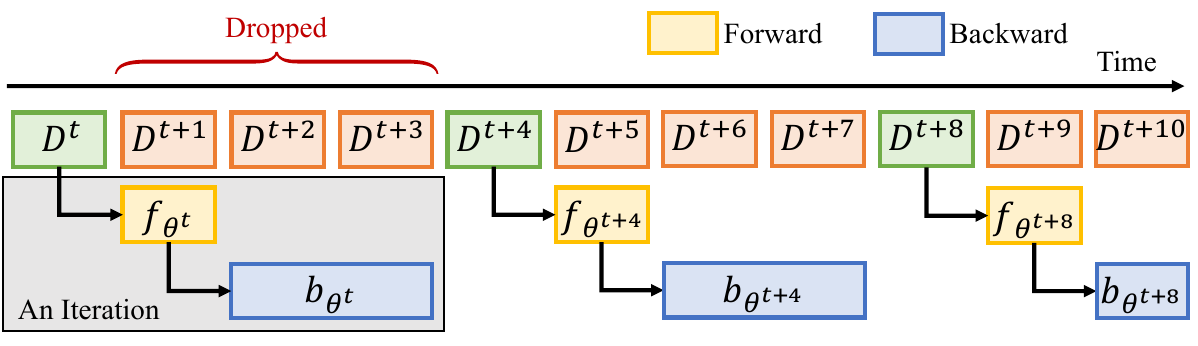}}
    \\
    \subfloat[Mini-batch online continual learning\label{fig:batched-sl}]{\includegraphics[width=0.45\textwidth]{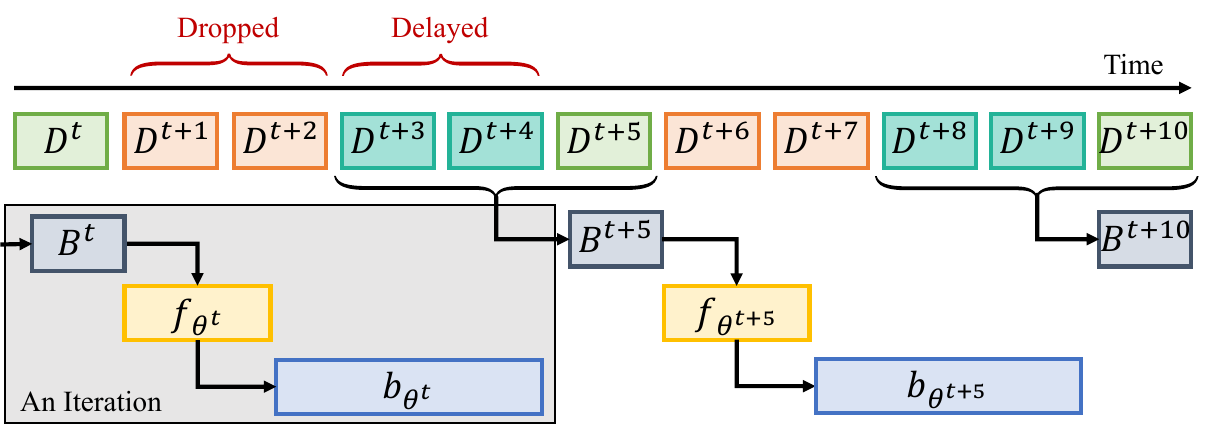}}
    \caption{(a) A typical online continual learning framework that trains the model as soon as the data arrives. (b) A mini-batch of data is processed for each iteration, reducing dropped data but introducing delayed data.}
\end{figure}

\textbf{Online Continual Learning: }
OCL is crucial in ML for adapting to continuously evolving data, primarily addressing catastrophic forgetting~\cite{chaudhry2019continual, aljundi2019online, li2017learning, aljundi2018memory} and concept drift in data streams under resource constraints~\cite{cai2021online, lin2021clear, fini2020online}.
Some methods~\cite{sahoo2017online, ghunaim2023real} reduce latency by immediately processing new data, as Fig.~\ref{fig:vanilla-sl} illustrated, but this often discards valuable information and hampers comprehensive learning.
To preserve data, several approaches~\cite{mirzasoleiman2020coresets, li2022camel, yang2023towards, zhou2023communication, zhou20253sfc} buffer it for later batch-training, as Fig.~\ref{fig:batched-sl} illustrated, which, while increasing latency and computational costs, especially with complex algorithms like coreset selection, seeks to avoid data skipping.

\textbf{Distributed and parallel ML frameworks: }
Distributed and parallel ML frameworks are essential for processing large data sets with minimal latency~\cite{paszke2019pytorch, shoeybi2019megatron, rasley2020deepspeed, sergeev2018horovod, moritz2018ray, zhou2024defta}.
For instance, Pytorch~\cite{paszke2019pytorch}, Megatron-LM~\cite{shoeybi2019megatron}, and DeepSpeed~\cite{rasley2020deepspeed} use GPipe~\cite{huang2019gpipe} and PipeDream-Flush~\cite{narayanan2021memory} to enable concurrent processing across model segments, enhancing real-time data management.\break 
Horovod~\cite{sergeev2018horovod} streamlines synchronized distributed training across TensorFlow~\cite{abadi2016tensorflow}, Keras~\cite{chollet2015}, and PyTorch, optimizing ML training on vast computing clusters. 
Conversely, Ray~\cite{moritz2018ray} supports large-scale asynchronous distributed training with custom gradient compensation algorithms~\cite{zheng2017asynchronous}.

\begin{figure}[t]
    \centering
    \includegraphics[width=1.0\linewidth]{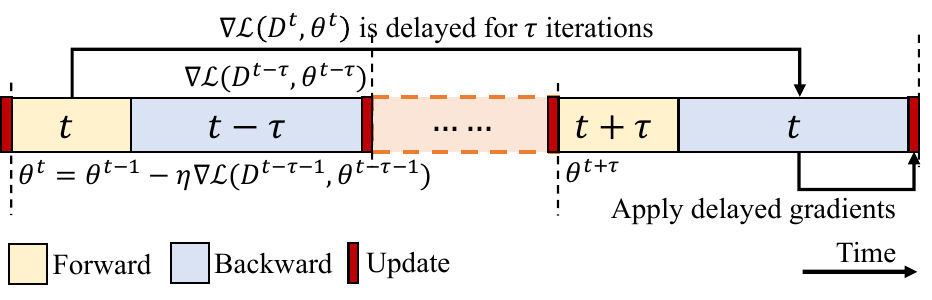}
    \caption{Due to asynchronous model updating, $\nabla \mathcal{L}(D^t, \theta^t)$ is obtained when $\theta^t$ has been updated to $\theta^{t+\tau}$}
    \label{fig:gradients-staleness}
\end{figure}

\textbf{Parallelization optimizations: }
Several strategies beyond data and model parallelism have been developed to optimize parallelization in ML.
Pipeline parallelism has seen notable advancements with the introduction of strategies like DAPPLE~\cite{fan2021dapple}, Pipedream~\cite{narayanan2019pipedream}, Zero-Bubble~\cite{qi2023zero}, and Hanayo~\cite{liu2023hanayo}, significantly boosting training efficiency.
Additionally, during asynchronous parallelization, the model will be inevitably updated by stale gradients, as Fig.~\ref{fig:gradients-staleness} illustrated, leading to low data management efficiency and performance degradation.
Strategies to counteract stale gradients in parallel processing include reducing step sizes for outdated gradients~\cite{hardy2017distributed, jiang2017heterogeneity, barkai2019gap} or using higher-order information~\cite{zheng2017asynchronous, zhou2021communication}.

\section{Notations}
\label{sec:notation}

\begin{table}[ht]
    \centering
    \caption{The glossary of notations}
    \resizebox{1.0\linewidth}{!}{
    \begin{tabular}{p{0.2\linewidth}p{0.8\linewidth}}
        \hline
        Notation & Implication \\
        \hline
        $\theta \in \Theta$ & Model parameters \\
        $\hat{L}$ & Number of layers in $\theta$ \\
        $\hat{t}_i^f$ & Time consumed for the $i$-th layer's forward pass \\
        $|\hat{w}_i|$ & Number of parameters in the $i$-th layer's \\
        $|\hat{a}_i|$ & Number of parameters for the $i$-th layer's output activations \\
        $L$ & Partition scheme, the $j$-th partition has $[L_j, L_{j+1})$ layers \\
        $P$ & Number of pipeline stages $|L| - 1$ \\
        $|w_j|$ & Number of parameters in the $j$-th stage $\sum_{i=L_j}^{L_{j+1} - 1}|\hat{w}_i|$ \\
        $|a_j|$ & Number of activations in the $j$-th stage $\sum_{i=L_j}^{L_{j+1} - 1}|\hat{a}_i|$ \\
        $t^f$ & Time consumed for the forward pass of a stage $\max_j \{\sum_{i=L_j}^{L_{j+1} - 1}|\hat{t}_i^f|\}$ \\
        $t^b$ & Time consumed for the backward pass of a stage $\max_j \{\sum_{i=L_j}^{L_{j+1} - 1}|\hat{t}_i^b|\}$ \\
        $t^d$ & Interval between data arrivals \\
        $C$ & Pipeline configuration \\
        $c_n^d$ & Processing delay of the $n$-th worker \\
        $c_n^r$ & Actication recomputation indicator of the $n$-th worker \\
        $c_{n,j}^a$ & Number of gradient accumulation steps of the $j$-th stage in the $n$-th worker \\
        $c_{n,j}^o$ & Number of back-propagation omission steps of the $j$-th stage in the $n$-th worker \\
        $D^t$ & Data arrived at timestamp $t$ \\
        $V_{D^t}$ & Data value of $D^t$ \\
        $\mathcal{R}_{\mathcal{A}}^T$ & Adaptation rate of a OCL framework $\mathcal{A}$ for $t \in [0, T]$ \\
        $\mathcal{M}_{\mathcal{A}}$ & Memory footprint of $\mathcal{A}$ \\
        $oacc_{\mathcal{A}}(t)$ & Online accuracy of $\mathcal{A}$ at timestamp $t$ \\
        $tacc_{\mathcal{A}}(t)$ & Test accuracy of $\mathcal{A}$ at timestamp $t$ \\
        $agm_{\mathcal{B}}(\mathcal{A}, t)$ & Online Accuracy Gain per unit of Memory of $\mathcal{A}$ over $\mathcal{B}$ at timestamp $t$ \\
        $tagm_{\mathcal{B}}(\mathcal{A}, t)$ & Test Accuracy Gain per unit of Memory of $\mathcal{A}$ over $\mathcal{B}$ at timestamp $t$ \\
        \hline
    \end{tabular}}
    \label{tab:glossory}
\end{table}

Let us assume the model $\theta$ in the hypothesis space $\Theta$ has $\hat{L}$ layers, with $\hat{t}_i^f$, $\hat{t}_i^b$, $|\hat{w}_i|$, $|\hat{a}_i|$ denoting the time consumed for the forward and backward pass, the size of the model parameters and the output activations of the $i$-th layer of the model, respectively.
The model is then partitioned into $P = |L| - 1$ pipeline stages, where $L$ represents the model partition scheme.
The $j$-th stage encompasses the $L_j$-th to the $(L_{j+1} - 1)$-th layer of the model, with the size of model parameters $|w_j| = \sum_{i=L_j}^{L_{j+1} - 1}|\hat{w}_i|$ and the size of activations $|a_j| = \sum_{i=L_j}^{L_{j+1} - 1}|\hat{a}_i|$.
Thus, the time consumed for the forward and backward pass of a stage in pipeline parallelism is $t^f = \max_j \{\sum_{i=L_j}^{L_{j+1} - 1}|\hat{t}_i^f|\}$ and $t^b = \max_j \{\sum_{i=L_j}^{L_{j+1} - 1}|\hat{t}_i^b|\}$, respectively.
Moreover, after partitioning, a pipeline configuration $C$ is determined based on the interval between data arrivals $t^d$ that contains $N$ worker configurations, where the $n$-th worker configuration specifies the processing delay $c_n^d$, the activation recomputation indicator $c_n^r$, and the gradient accumulation and back-propagation omission steps of its $j$-th stage $c_{n,j}^a$, $c_{n,j}^o$, respectively.

\section{Adaptation Rate and Memory Reduction for S1-S4}

In Sec.~\ref{sec:ics}, \textbf{T1}-\textbf{T4} are progressively employed to reduce $\mathcal{R}_{F}^T$ and $\mathcal{M}_{F}$.

\textbf{S1. Deploy T1 for all workers: } 
By setting $c_n^r=1$ for all workers, the data processing time increases.
Specifically, for the $n$-th worker, setting $c_n^r=1$ will respectively reduce $\mathcal{R}_{F}^T$ and $\mathcal{M}_{F}$ by:
{\small  
  \setlength{\abovedisplayskip}{6pt}
  \setlength{\belowdisplayskip}{\abovedisplayskip}
  \setlength{\abovedisplayshortskip}{0pt}
  \setlength{\belowdisplayshortskip}{3pt}
\begin{align}
     &\Delta \mathcal{R}_{F}^T = \sum_{i=0}^{P-1} \frac{|w_i|}{\sum_{j=0}^{P-1} (|w_j|)} \frac{1}{c_{n,i}^a} \sum_{j=0}^{c_{n,i}^a - 1} (B_{i,j} - C_{i,j}), \nonumber\\
     &\mathrm{where}~B_{i,j} = \frac{e^{-c((2P + 2j - i) t^f + (P-i+j) t^b)} V_D}{LCM(\{c_{n,k}^o + 1 | k \in [i,P-1]\}) (2 t^f + t^b)},\nonumber \\
     &C_{i,j} = \frac{e^{-c((P + j) t^f + (P-i+j) t^b)} V_D}{LCM(\{c_{n,k}^o + 1 | k \in [i,P-1]\}) (t^f + t^b)}, \nonumber \\
     &\Delta \mathcal{M}_{F} = - \sum_{i=0}^{P-1} (1 + \lceil \frac{P - i - 1}{c_{n,i}^a} \rceil - c_{n,i}^o) \sum_{l=L_i + 1}^{L_{i+1}-1} |\hat{a}_l|.
\label{eq:dsvre-ar}
\end{align}
}

\textbf{S2. Deploy T2 for the $j$-th stage in the $n$-th worker: }
If $c_{n,j}^o=0$, increasing $c_{n,j}^a$ by $\Delta c_{n,j}^a = \lceil \frac{P - j - 1}{\lceil (P - j - 1) / c_{n,j}^a \rceil - 1} \rceil - c_{n,j}^a$ will lead to a reduced frequency of model parameter updates.
Here, the value of $\Delta c_{n,j}^a$ is determined to prevent $\Delta_{c_{n,j}^a \rightarrow c_{n,j}^a + 1} \mathcal{M}_{F} = 0$ due to the ceiling function.
Consequently, $\mathcal{R}_{F}^T$ and $\mathcal{M}_{F}$ will be respectively decreased by:
{\small  
  \setlength{\abovedisplayskip}{6pt}
  \setlength{\belowdisplayskip}{\abovedisplayskip}
  \setlength{\abovedisplayshortskip}{0pt}
  \setlength{\belowdisplayshortskip}{3pt}
\begin{align}
    &\Delta \mathcal{R}_{F}^T = \frac{|w_j|}{\sum_{k=0}^{P-1} (|w_k|)} (\frac{\sum_{k=c_{n,j}^a}^{c_{n,j}^a + \Delta c_{n,j}^a - 1} A_{j,k}}{c_{n,j}^a + \Delta c_{n,j}^a} - \frac{\Delta c_{n,j}^a \sum_{k=0}^{c_{n,j}^a - 1} A_{j,k}}{(c_{n,j}^a + \Delta c_{n,j}^a)c_{n,j}^a}), \nonumber \\
    &\Delta \mathcal{M}_{F} = -(|w_j| + |a_j| - c_n^r \sum_{l=L_j + 1}^{L_{j+1}-1} |\hat{a}_l|).
\label{eq:dsvre-ga}
\end{align}
}

\textbf{S3. Deploy T3 For the $j$-th stage in the $n$-th worker: } 
If $\Delta c_{n,j}^a = +\infty$, setting $c_{n,j}^a = 1$ and $c_{n,j}^o=P - 1 - j$ will completely eliminate the need for the $j$-th stage in the $n$-th worker to store additional model parameters by bypassing any backward pass that requires previous model parameters.
Consequently, $\mathcal{R}_{F}^T$ and $\mathcal{M}_{F}$ will be respectively reduced by:
{\small  
  \setlength{\abovedisplayskip}{6pt}
  \setlength{\belowdisplayskip}{\abovedisplayskip}
  \setlength{\abovedisplayshortskip}{0pt}
  \setlength{\belowdisplayshortskip}{3pt}
\begin{align}
    &\Delta \mathcal{R}_{F}^T = \sum_{i=0}^{P-1-j} \frac{|w_i|}{\sum_{k=0}^{P-1} (|w_j|)} \frac{\sum_{j=0}^{c_{n,i}^a - 1} (E_{i,j} - A_{i,j})}{c_{n,i}^a}, \mathrm{where}~E_{i,j} = \nonumber \\
    &\frac{e^{-c((P + j) t^f + (P-i+j) t^b + c_n^r (P-i+j) t^f)} V_D}{LCM(LCM(\{c_{n,k}^o + 1 | k \in [i,P-1]\}), P - j) (t^f + t^b + c_n^r t^f)}, \nonumber \\
    &\Delta \mathcal{M}_{F} = - \lceil \frac{P - j - 1}{c_{n,j}^a} \rceil (|w_j| + |a_j| - c_n^r \sum_{l=L_j + 1}^{L_{j+1}-1} |\hat{a}_l|).
\label{eq:dsvre-bpo}
\end{align}
}

\textbf{S4. Deploy T4 for the $n$-th worker: }
If $c_{n,j}^o \neq 0$ for all $j \in [0, p-1)$, shutting down the $n$-th worker will lead to a respective decrease in $\mathcal{R}_{F}^T$ and $\mathcal{M}_{F}$ by:
{\small  
  \setlength{\abovedisplayskip}{6pt}
  \setlength{\belowdisplayskip}{\abovedisplayskip}
  \setlength{\abovedisplayshortskip}{0pt}
  \setlength{\belowdisplayshortskip}{3pt}
\begin{align}
    &\Delta \mathcal{R}_{F}^T = - \sum_{i=0}^{P-1} \frac{|w_i|}{\sum_{j=0}^{P-1} (|w_j|)} \frac{1}{c_{n,i}^a} \sum_{j=0}^{c_{n,i}^a - 1} A_{i,j}, \nonumber\\
    &\mathrm{where}~A_{i,j} = \frac{e^{-c((P + j) t^f + (P-i+j) t^b + c_n^r (P-i+j) t^f)} V_D}{LCM(\{c_{n,k}^o + 1 | k \in [i,P-1]\}) (t^f + t^b + c_n^r t^f)}, \nonumber \\
    &\Delta \mathcal{M}_{F} = - \sum_{i=0}^{P-1} (1 + \lceil \frac{P - i - 1}{c_{n,i}^a} \rceil - c_{n,i}^o) (|w_i| + |a_i| - c_n^r \sum_{l=L_i + 1}^{L_{i+1}-1} |\hat{a}_l|).
\label{eq:dsvre-d}
\end{align}
}

\section{Algorithm Details}
\label{sec:algorithm}

\begin{algorithm}[htbp]
    \DontPrintSemicolon
    \SetKwInput{KwInput}{Input}
    \SetKwInput{KwOutput}{Output}
    \SetKwProg{compensate}{Function \emph{compensate}}{}{end}
    \KwInput{$\nabla \mathcal{L}(D^{t-1}; \theta^{t-1})$; $\theta^{t-1}, \theta^{t}, \ldots, \theta^{t+\tau-1}$; $\alpha$; $\eta$ and $\eta_\lambda$; $\lambda^0$;}
    \KwOutput{$\theta^{t+\tau}$;}
    \compensate{($\nabla \mathcal{L}(D^{t-1}; \theta^{t-1})$, $\theta^{t-1}$, $\theta^{t}$, $\ldots$, $\theta^{t+\tau-1}$)}{
        \tcp*[l]{initialization}\
        $\lambda = \lambda^0$; $v_r = v_a = 0$\

        \uIf{$\eta_\lambda > 0$}{
            \tcp*[l]{update $\lambda$}\
            $\Delta v_r = (1 - \alpha) (\nabla \mathcal{L}(D^{t-1}; \theta^{t-1}) - v_r)$\
    
            $\lambda = \lambda - \eta_\lambda \nabla_\lambda ||\Delta v_r - \lambda v_a||^2$\
    
            $v_r = \alpha v_r + (1 - \alpha) \nabla \mathcal{L}(D^{t-1}; \theta^{t-1})$\
            
            $v_a = \alpha v_a + (1 - \alpha) (\nabla \mathcal{L}(D^{t-1}; \theta^{t-1}) \odot \nabla \mathcal{L}(D^{t-1}; \theta^{t-1}) \odot (\theta^t - \theta^{t - 1}))$\
        }

        \tcp*[l]{iterative approximation}\
        \ForAll{$i$ in $0,...,\tau-1$}{
            $\nabla \mathcal{L}(D^{t-1}; \theta^{t+i}) = A_\mathcal{I}(\nabla \mathcal{L}(D^{t-1}; \theta^{t+i-1}), \theta^{t+i}, \theta^{t+i-1})$ \
        }

        \tcp*[l]{model update}\
        $\theta^{t+k} = \theta^{t+\tau-1} - \eta \nabla \mathcal{L}(D^{t-1}; \theta^{t+\tau-1})$\
        
        \Return{$\theta^{t+\tau}$}\
    }
    \caption{Iterative Gradients Compensation with $\lambda$}
    \label{alg:iterative-gradients-compensation}
\end{algorithm}

Algorithm~\ref{alg:iterative-gradients-compensation} initially updates $\lambda$ for lower approximation errors (Line 3-7).
Then, the algorithm iteratively reduces the staleness of gradients step by step, given the hyper-parameter $\lambda$ (Line 8-9). 
Afterward, the model parameter $\theta^{t+\tau-1}$ is updated by the delay-compensated gradient $\nabla \mathcal{L}(D^{t-1}; \theta^{t+\tau-1})$ using gradient descent algorithm (Line 10).

\begin{algorithm}[htbp]
    \DontPrintSemicolon
    \SetKwInput{KwInput}{Input}
    \SetKwInput{KwOutput}{Output}
    \SetKwProg{itersearch}{Function \emph{itersearch}}{}{end}
    \SetKwProg{search}{Function \emph{search}}{}{end}
    \KwInput{$t^d$; $t^f$; $t^b$; $L$; $M$;}
    \KwOutput{$C = \{c_n^r, c_n^d, c_{n,j}^a, c_{n,j}^o\}$; $\mathcal{R}_{F}^T$;}
    \itersearch{($t^d$, $t^f + t^b$, $c^r$, $L$, $M$)}{
        \tcp*[l]{initialization}\
        $N = \lceil (t^f + t^b + c^r t^f)/t^d \rceil$\

        $c_n^d = n$; $c_n^r = c^r$; $c_{n,j}^a = 1$; $c_{n,j}^o = 0$\
        
        \tcp*[l]{iterative search}\
        Calculate $\mathcal{R}_{F}^T$ and $\mathcal{M}_{F}$\
        
        \While{$\mathcal{M}_{F} > M$}{
            \ForAll{$n$ in $0, \dots, N - 1$}{
                \ForAll{$p$ in all \{\textbf{S2, S3, S4}\}}{
                    Calculate $\Delta_p \mathcal{M}_{F}/\Delta_p \mathcal{R}_{F}^T$
                }
            }
            
            $p^* = \arg \max_p(\Delta_p \mathcal{M}_{F}/\Delta_p \mathcal{R}_{F}^T)$ \
            
            Update $c_n^r$, $c_n^d$, $c_{n,j}^a$, $c_{n,j}^o, \mathcal{R}_{F}^T, \mathcal{M}_{F}$ based on $p^*$\

        }

        \Return{$C = \{c_n^r, c_n^d, c_{n,j}^a, c_{n,j}^o\}, \mathcal{R}_{F}^T$}\
    }

    \search{($t^d$, $t^f + t^b$, $L$, $M$)}{
        $C_0, \mathcal{R}_0 = itersearch(t^d, t^f + t^b, 0, L, M)$\
        
        $C_1, \mathcal{R}_1 = itersearch(t^d, t^f + t^b, 1, L, M)$\

        \Return{$C_i, \mathcal{R}_i$ where $i = \arg\max_i \{\mathcal{R}_i\}$}
    }
    \caption{Iterative Configuration Search}
    \label{alg:iter-conf-search}
\end{algorithm}

Algorithm~\ref{alg:iter-conf-search} begins by initializing $N$, $c_n^d$, $c_n^r$, $c_{n,j}^a$ and $c_{n,j}^o$ using the given parameters $P$, $t^d$, $t^f$, $c^r$ and $t^b$ (Lines 2-3).
Then, the initial $\mathcal{R}_{F}^T$ and $\mathcal{M}_{F}$ are computed using Eq.~\ref{eq:ferret-svre} (Line 4).
The algorithm then repeatedly evaluate $\Delta_p \mathcal{M}_{F}$ and $\Delta_p \mathcal{R}_{F}^T$ based on the aforementioned deployments $\{$\textbf{S2}, \textbf{S3}, \textbf{S4}$\}$, applying the one that maximizes $\Delta_p \mathcal{M}_{F}/\Delta_p \mathcal{R}_{F}^T$ until $\mathcal{M}_{F} \leq M$ (Lines 5-10).
Finally, \textbf{S1} is evaluated separately in Lines 13-14 since it is applied to all workers and impacts $t^b$ and $N$.

\begin{algorithm}[htbp]
    \DontPrintSemicolon
    \SetKwInput{KwInput}{Input}
    \SetKwInput{KwOutput}{Output}
    \SetKwProg{plan}{Function \emph{plan}}{}{end}
    \KwInput{$t^d$; $\theta$; $M$; $profile(\cdot)$; $search(\cdot)$;}
    \KwOutput{$L^*$ and $C^*$ that maximizes Eq.~\ref{eq:objective};}
    \plan{($t^d$, $\theta$, $M$)}{
        $\hat{L}, \hat{t}_i^f, \hat{t}_i^b, |\hat{w}_i|, |\hat{a}_i| = profile(\theta)$; $S = \varnothing$\\

        \tcp*[l]{get all consumed time for a stage}\
        \ForAll{$k$ in $0, \dots, \hat{L} - 1$}{
            
            \ForAll{$i$ in $0, \dots, \hat{L} - 1 - k$}{
                
                $t^c = \sum_{j=i}^{i + k}(\hat{t}_j^f + \hat{t}_j^f)$\
                
                \If{$(t^c < \max_{j \in [0, L-1]}(\hat{t}_j^f + \hat{t}_j^b)$}{
                    $S = S \cup t^c$\
                }
            }
        }

        $S = sorted(S)$

        \tcp*[l]{Search for the optimal scheme}\
        $L^* = []; {t^c}^* = \mathcal{R}^* = 0; C^* = \varnothing$\
        
        \ForAll{$t^c$ in $S$}{

            $L = [0]; t^{sum} = 0$\

            \ForAll{$i$ in $0, \dots, \hat{L} - 1$}{
                
                $t^{sum} = t^{sum} + \hat{t}_i^f + \hat{t}_i^b$

                \If{$t^{sum} > t^c$}{
                
                    $L.append(i); t^{sum} = \hat{t}_i^f + \hat{t}_i^b$\
                }
            }

            $L.append(\hat{L})$\

            $C, \mathcal{R} = search(t^d, t^c, L, M)$\

            \If{$\mathcal{R} > \mathcal{R}^*$}{
                $\mathcal{R}^* = \mathcal{R}; {t^c}^* = t^c; L^* = L; C^*=C;$\
                
            }
        }
        
        \Return{$L^*$, $C^*$}\
    }
    \caption{Brute-force planning}
    \label{alg:brute-force-planning}
\end{algorithm}

Algorithm~\ref{alg:brute-force-planning} starts by listing all possible $t^c$ for a stage, where $profile(\cdot)$ measures the number of layers and their respective statistics. (Line 3-8).
For each $t^c$, a model partition scheme $L$ is then constructed (Line 11-16).
Finally, the optimal $C$ and $\mathcal{R}$ under the formed $L$ are retrieved using the $search(\cdot)$ in Alg.~\ref{alg:iter-conf-search} (Line 17).
By comparing $\mathcal{R}$ of different $L$, $L^*$ and $C^*$ can be obtained (Line 18-19).

\section{Implementation Details}
\label{sec:implementation}

\begin{table}[htbp]
  \centering
  \caption{Statistics of raw datasets.}
  \resizebox{1.0\linewidth}{!}{%
    \begin{tabular}{lrrr}
    \toprule
    Dataset & \multicolumn{1}{l}{\# Stream Data} & \multicolumn{1}{l}{\# Features} & \multicolumn{1}{l}{\# Classes} \\
    \midrule
    MNIST~\cite{deng2012mnist} & 60,000 & 784   & 10 \\
    FMNIST~\cite{xiao2017fashion} & 60,000 & 784   & 10 \\
    EMNIST~\cite{cohen2017emnist} & 697,932 & 784   & 62 \\
    CIFAR10~\cite{krizhevsky2009learning} & 50,000 & 3,072 & 10 \\
    CIFAR100~\cite{krizhevsky2009learning} & 50,000 & 3,072 & 100 \\
    SVHN~\cite{netzer2011reading}  & 604,388 & 3,072 & 10 \\
    Tiny-ImageNet~\cite{le2015tiny} & 100,000 & 12,288 & 200 \\
    CORe50~\cite{lomonaco2017core50} & 119,894 & 49,152 & 50 \\
    CORe50-iid & 119,894 & 49,152 & 50 \\
    Split-MNIST & 60,000 & 784   & 10 \\
    Split-FMNIST & 60,000 & 784   & 10 \\
    Split-CIFAR10 & 50,000 & 3,072 & 10 \\
    Split-CIFAR100 & 50,000 & 3,072 & 100 \\
    Split-SVHN & 73,257 & 3,072 & 10 \\
    Split-Tiny-ImageNet & 100,000 & 12,288 & 200 \\
    Covertype~\cite{misc_covertype_31} & 464,809 & 54    & 7 \\
    CLEAR10~\cite{lin2021clear} & 330,000 & 562,500 & 11 \\
    CLEAR100~\cite{lin2021clear} & 1,209,197 & 562,500 & 101 \\
    \bottomrule
    \end{tabular}%
    }
  \label{tab:dataset-statistics}%
\end{table}%

\textbf{Datasets:} To align with our computational resources, images in Tiny-ImageNet~\cite{le2015tiny} and CORe50~\cite{lomonaco2017core50} are resized to $32\times32$, while images in CLEAR10 and CLEAR100~\cite{lin2021clear} are resized to $224\times224$.
Additionally, split datasets (\textit{i.e.}, Split-MNIST, Split-CIFAR10, etc.) are partitioned into 5 tasks to simulate a class-incremental setting~\cite{buzzega2020dark, de2021continual, li2022camel}.
CORe50-iid is a shuffled version of the CORe50 dataset.

\textbf{The rationale behind our evaluation metrics:} To comprehensively evaluate both performance and memory footprint of given OCL frameworks $\mathcal{A}$ and $\mathcal{B}$, it is natural to consider the ratio of their differences:
\begin{equation}
    metric_{c} = \frac{metric_{\mathcal{A}} - metric_{\mathcal{B}}}{\log(M_{\mathcal{A}}) - \log(M_{\mathcal{B}})},
\label{eq:metric-rationale}
\end{equation}
where $metric_{c}$ is the comprehensive metric that shows the performance improvement per memory increment, and $metric_{\mathcal{A}}$ and $metric_{\mathcal{B}}$ are the arbitrary performance metrics of $\mathcal{A}$ and $\mathcal{B}$, respectively.
It is important to note that the logarithm function is employed to account for the diminishing returns associated with memory increments.
Then, Eq.~\ref{eq:metric-rationale} can be further represented as:
\begin{align}
    metric_{c} &\sim \frac{metric_{\mathcal{A}} - metric_{\mathcal{B}}}{\log(M_{\mathcal{A}} / M_{\mathcal{B}})} \\
    &\sim \frac{\exp(metric_{\mathcal{A}} - metric_{\mathcal{B}})}{M_{\mathcal{A}} / M_{\mathcal{B}}} \\
    &\sim \log(\frac{\exp(metric_{\mathcal{A}} - metric_{\mathcal{B}})}{M_{\mathcal{A}} / M_{\mathcal{B}}}),
\end{align}
which is the same form as Eq.~\ref{eq:metric-tagm} and Eq.~\ref{eq:metric-agm}.

\textbf{Hyper-parameters:} All experiments are conducted on a server with 32 Intel(R) Xeon(R) CPU E5-2620 v42.10GHz CPUs, 8 NVIDIA TITAN Xp GPUs and 64 GB memory.
The learning rate is 1e-3, the interval between data arrivals is set to maximal time consumed for the forward pass of a layer in the model, \textit{i.e.}, $t^d = \max_i \{ \hat{t}_i^f \}$, and the replay buffer size for all OCL algorithms is 5e3.
For Iter-Fisher, $\lambda$ = 0.2, $\mu$ = 2e-6, while for the other compensation algorithms, $\lambda$ is manually tuned for each dataset.
Additionally, $L^*$ and $C^*$ are pre-determined and shared for all pipeline parallelism strategies using tuned $c$ and $M$ for each dataset.

\section{Additional Evaluation Results}
\label{sec:additional-results}

In this section, we present additional results for the proposed Ferret and the baseline methods, employing the standard Online Accuracy and Test Accuracy metrics.
These metrics are widely recognized as the standard measure in the field of OCL.
Notably, The following results do not account for the memory footprint during training, which is the main reason why they are not included in the main paper.
  
\begin{table*}[htbp]
    \centering
    \caption{Online Accuracy of different algorithms. "M-", "M", "M+" refer to the ferret method with minimal, medium and maximal memory footprint, respectively.}
    \resizebox{1.0\linewidth}{!}{%
      \begin{tabular}{l|r|rrrr|rrr}
      \toprule
      Setting & \cellcolor{gray!20}Oracle  & 1-Skip  & Random-$N$ & Last-$N$ & Camel & Ferret$_{\mathrm{M-}}$ & Ferret$_{\mathrm{M}}$ & Ferret$_{\mathrm{M+}}$ \\
      \midrule
      MNIST/MNISTNet & \cellcolor{gray!20}81.14$_{\pm1.43}$ & 18.24$_{\pm3.05}$ & 18.48$_{\pm2.65}$ & 18.87$_{\pm3.17}$ & 17.82$_{\pm2.7}$& 30.16$_{\pm4.57}$ & \underline{56.25}$_{\pm3.69}$ & \textbf{80.98}$_{\pm1.44}$ \\
      FMNIST/MNISTNet & \cellcolor{gray!20}65.94$_{\pm0.64}$ & 21.39$_{\pm2.71}$ & 21.9$_{\pm1.66}$ & 22.02$_{\pm1.57}$ & 21.22$_{\pm2.12}$ & 34.76$_{\pm2.89}$ & \underline{51.19}$_{\pm1.72}$ & \textbf{65.78}$_{\pm0.66}$ \\
      EMNIST/MNISTNet & \cellcolor{gray!20}75.91$_{\pm0.14}$ & 45.98$_{\pm1.23}$ & 51.64$_{\pm1.14}$ & 51.82$_{\pm1.15}$ & 50.73$_{\pm1}$ & 55.33$_{\pm1.04}$ & \underline{66.8}$_{\pm0.31}$ & \textbf{75.9}$_{\pm0.15}$ \\
      Cifar10/ConvNet & \cellcolor{gray!20}51.84$_{\pm0.04}$ & 27.5$_{\pm0.19}$ & 40.09$_{\pm0.26}$ & 40.24$_{\pm0.21}$ & 40.06$_{\pm0.11}$ & 34.25$_{\pm0.56}$ & \underline{42.46}$_{\pm0.18}$ & \textbf{51.69}$_{\pm0.36}$ \\
      Cifar100/ConvNet & \cellcolor{gray!20}14.56$_{\pm0.03}$ & 2.49$_{\pm0.01}$ & 5.81$_{\pm0.16}$ & 5.92$_{\pm0.13}$ & 5.75$_{\pm0.17}$ & 5.66$_{\pm0.08}$ & \underline{9.15}$_{\pm0.06}$ & \textbf{14.89}$_{\pm0.12}$ \\
      SVHN/ConvNet & \cellcolor{gray!20}81.59$_{\pm0.04}$ & 46.11$_{\pm0.52}$ & 64.07$_{\pm0.68}$ & 64.52$_{\pm0.61}$ & 64.86$_{\pm0.73}$ & 56.98$_{\pm0.75}$ & \underline{73.29}$_{\pm0.25}$ & \textbf{81.6}$_{\pm0.2}$ \\
      TinyImagenet/ConvNet & \cellcolor{gray!20}5.65$_{\pm0.19}$ & 0.75$_{\pm0.02}$ & 1.6$_{\pm0.12}$ & 1.63$_{\pm0.1}$ & 1.62$_{\pm0.13}$ & 1.51$_{\pm0.08}$ & \underline{2.67}$_{\pm0.07}$ & \textbf{5.61}$_{\pm0.17}$ \\
      CORe50/ConvNet & \cellcolor{gray!20}81.59$_{\pm0.12}$ & 21.69$_{\pm1.01}$ & 51.25$_{\pm0.54}$ & 51.59$_{\pm0.66}$ & 48.81$_{\pm0.74}$ & 42.05$_{\pm0.97}$ & \underline{63.68}$_{\pm0.56}$ & \textbf{80.46}$_{\pm0.22}$ \\
      CORe50-iid/ConvNet & \cellcolor{gray!20}63.8$_{\pm0.34}$ & 19.49$_{\pm6.82}$ & 27.74$_{\pm6.39}$ & 34.34$_{\pm0.67}$ & 33.14$_{\pm1.24}$ & 27.11$_{\pm1.09}$ & \underline{44.89}$_{\pm0.66}$ & \textbf{63.23}$_{\pm0.34}$ \\
      SplitMNIST/MNISTNet & \cellcolor{gray!20}94.96$_{\pm0.08}$ & 53.04$_{\pm1.82}$ & 59.66$_{\pm2.7}$ & 59.72$_{\pm2.99}$ & 61.85$_{\pm2.73}$ & 66.81$_{\pm3.5}$ & \underline{87.11}$_{\pm0.61}$ & \textbf{94.4}$_{\pm0.2}$ \\
      SplitFMNIST/MNISTNet & \cellcolor{gray!20}94.98$_{\pm0.29}$ & 68.92$_{\pm3.68}$ & 73.67$_{\pm4.23}$ & 73.57$_{\pm4.27}$ & 74.66$_{\pm4.17}$ & 81.14$_{\pm2.68}$ & \underline{91.08}$_{\pm0.6}$ & \textbf{94.7}$_{\pm0.29}$ \\
      SplitCifar10/ConvNet & \cellcolor{gray!20}84.1$_{\pm0.07}$ & 66.84$_{\pm0.21}$ & 75.6$_{\pm0.13}$ & 75.73$_{\pm0.32}$ & 75.76$_{\pm0.07}$ & 72.89$_{\pm0.24}$ & \underline{78.64}$_{\pm0.31}$ & \textbf{83.55}$_{\pm0.09}$ \\
      SplitCifar100/ConvNet & \cellcolor{gray!20}33.52$_{\pm0.24}$ & 9.33$_{\pm0.2}$& 17.32$_{\pm0.19}$ & 17.44$_{\pm0.27}$ & 17.17$_{\pm0.5}$ & 17.01$_{\pm0.16}$ & \underline{24.1}$_{\pm0.2}$ & \textbf{33.72}$_{\pm0.14}$ \\
      SplitSVHN/ConvNet & \cellcolor{gray!20}94.83$_{\pm0.02}$ & 79.89$_{\pm0.77}$ & 88.31$_{\pm0.48}$ & 88.32$_{\pm0.55}$ & 88.29$_{\pm0.48}$ & 85.58$_{\pm0.49}$ & \underline{92.06}$_{\pm0.13}$ & \textbf{94.74}$_{\pm0.05}$ \\
      SplitTinyImagenet/ConvNet & \cellcolor{gray!20}5.72$_{\pm0.27}$ & 0.8$_{\pm0.06}$ & 1.59$_{\pm0.13}$ & 1.67$_{\pm0.1}$ & 1.55$_{\pm0.06}$ & 1.53$_{\pm0.08}$ & \underline{2.9}$_{\pm0.07}$ & \textbf{5.66}$_{\pm0.2}$ \\
      CLEAR10/ResNet & \cellcolor{gray!20}96.4$_{\pm0.02}$ & 72.52$_{\pm0.15}$ & 91.42$_{\pm0.17}$ & 91.63$_{\pm0.13}$ & 66.71$_{\pm24.22}$ & 78.02$_{\pm0.02}$ & \underline{90.96}$_{\pm0.03}$ & \textbf{96.32}$_{\pm0.04}$ \\
      CLEAR10/MobileNet & \cellcolor{gray!20}76.89$_{\pm0.51}$ & 30.02$_{\pm0.48}$ & 58.75$_{\pm0.67}$ & 59.22$_{\pm0.4}$ & 58.88$_{\pm0.33}$ & 25.36$_{\pm0.78}$ & \underline{64.4}$_{\pm0.82}$ & \textbf{75.18}$_{\pm0.39}$ \\
      CLEAR100/ResNet & \cellcolor{gray!20}89.08$_{\pm0.06}$ & 39.08$_{\pm0.96}$ & 74.9$_{\pm0.16}$ & 75.29$_{\pm0.1}$ & 73.07$_{\pm0.11}$ & 56.24$_{\pm0.06}$ & \underline{75.53}$_{\pm0.31}$ & \textbf{89.54}$_{\pm0.37}$ \\
      CLEAR100/MobileNet & \cellcolor{gray!20}61$_{\pm2.49}$ & 6.85$_{\pm0.12}$ & 29.48$_{\pm0.69}$ & 30$_{\pm0.45}$ & 28.48$_{\pm0.25}$ & 8.69$_{\pm0.25}$ & \underline{43.81}$_{\pm0.97}$ & \textbf{60.29}$_{\pm1.44}$ \\
      Covertype/MLP & \cellcolor{gray!20}80.9$_{\pm0.69}$ & 63.25$_{\pm0.83}$ & 60.59$_{\pm1.52}$ & 60.66$_{\pm1.53}$ & 60.57$_{\pm1.49}$ & 64.94$_{\pm0.92}$ & \underline{67.63}$_{\pm0.23}$ & \textbf{72.95}$_{\pm0.17}$ \\
      \bottomrule
      \end{tabular}%
    }
    \label{tab:add-online-acc}%
\end{table*}%

Table~\ref{tab:add-online-acc} shows the online accuracy of different OCL frameworks on various datasets.
Thanks to pipeline parallelism, the model parameters can be updated more frequently, leading to a higher online accuracy even for Ferret$_{\mathrm{M-}}$.
Ferret$_{\mathrm{M+}}$, which has the highest memory footprint, achieves the best online accuracy on all datasets.
The results demonstrate that Ferret can effectively utilize memory resources to improve the online accuracy of the model.

\begin{table*}[htbp]
    \centering
    \captionof{table}{Online Accuracy and Test Accuracy of different integrated OCL algorithms on CORe50/ConvNet. Camel has its dedicated component to mitigate catastrophic forgetting and cannot be integrated with various OCL algorithm.}
    \resizebox{1.0\linewidth}{!}{%
        \begin{tabular}{rl|l|llll|lll}
      \toprule
            & Metric & \cellcolor{gray!20}Oracle  & 1-Skip  & Random-$N$ & Last-$N$ & Camel & Ferret$_{\mathrm{M-}}$ & Ferret$_{\mathrm{M}}$ & Ferret$_{\mathrm{M+}}$ \\
      \midrule
      \multicolumn{1}{l}{Vanilla} & $oacc$   & \cellcolor{gray!20}81.59$_{\pm0.12}$ & 21.69$_{\pm1.01}$ & 51.25$_{\pm0.54}$ & 51.59$_{\pm0.66}$ & 48.81$_{\pm0.74}$ & 42.05$_{\pm0.97}$ & \underline{63.68}$_{\pm0.56}$ & \textbf{80.46}$_{\pm0.22}$ \\
            & $tacc$     & \cellcolor{gray!20}15.68$_{\pm0.72}$ & 10.24$_{\pm0.82}$ & 14.35$_{\pm1.08}$ & 14.11$_{\pm0.42}$ & \underline{15.28}$_{\pm0.39}$& 12.02$_{\pm0.36}$ & 15.07$_{\pm0.68}$ & \textbf{15.83}$_{\pm0.9}$ \\
      \multicolumn{1}{l}{ER~\cite{chaudhry2019continual}} & $oacc$     & \cellcolor{gray!20}79.84$_{\pm0.09}$ & 24.51$_{\pm0.69}$ & 42.59$_{\pm0.41}$ & 43.2$_{\pm0.41}$ &  -    & 40.9$_{\pm0.83}$ & \underline{61.57}$_{\pm0.47}$ & \textbf{78.68}$_{\pm0.11}$ \\
            & $tacc$     & \cellcolor{gray!20}24.54$_{\pm0.49}$ & 14.91$_{\pm0.52}$ & 19.4$_{\pm0.68}$ & 20.32$_{\pm0.14}$ & -     & 16.8$_{\pm0.72}$ & \underline{22.06}$_{\pm0.11}$ & \textbf{24.32}$_{\pm0.81}$ \\
      \multicolumn{1}{l}{MIR~\cite{aljundi2019online}} & $oacc$     & \cellcolor{gray!20}79.84$_{\pm0.09}$ & 24.51$_{\pm0.69}$ & 42.53$_{\pm0.49}$ & 43.1$_{\pm0.35}$ & -     & 40.9$_{\pm0.83}$ & \underline{61.57}$_{\pm0.47}$ & \textbf{78.68}$_{\pm0.11}$ \\
            & $tacc$     & \cellcolor{gray!20}24.54$_{\pm0.49}$ & 14.91$_{\pm0.52}$ & 19.78$_{\pm0.66}$ & 20$_{\pm0.51}$ & -     & 16.8$_{\pm0.72}$ & \underline{22.06}$_{\pm0.11}$ & \textbf{24.32}$_{\pm0.81}$ \\
      \multicolumn{1}{l}{LWF~\cite{li2017learning}} & $oacc$     & \cellcolor{gray!20}81.6$_{\pm0.13}$ & 21.7$_{\pm1.02}$ & 51.25$_{\pm0.54}$ & 51.61$_{\pm0.67}$ & -     & 42.11$_{\pm0.85}$ & \underline{63.55}$_{\pm0.49}$ & \textbf{80.56}$_{\pm0.28}$ \\
            & $tacc$     & \cellcolor{gray!20}15.68$_{\pm0.72}$ & 10.24$_{\pm0.82}$ & 14.35$_{\pm1.08}$ & 14.11$_{\pm0.42}$ & -     & 12$_{\pm0.36}$ & \underline{15.07}$_{\pm0.68}$ & \textbf{15.83}$_{\pm0.9}$ \\
      \multicolumn{1}{l}{MAS~\cite{aljundi2018memory}} & $oacc$     & \cellcolor{gray!20}81.66$_{\pm0.18}$ & 22.11$_{\pm0.7}$ & 51.19$_{\pm0.42}$ & 51.62$_{\pm0.75}$ & -     & 41.8$_{\pm0.86}$ & \underline{63.57}$_{\pm0.59}$ & \textbf{80.48}$_{\pm0.21}$ \\
            & $tacc$     & \cellcolor{gray!20}16.76$_{\pm0.53}$ & 10.54$_{\pm0.59}$ & 13.75$_{\pm0.68}$ & 13.98$_{\pm1.11}$ & -     & 11.56$_{\pm0.09}$ & \underline{14.87}$_{\pm0.73}$ & \textbf{16.48}$_{\pm0.98}$ \\
      \bottomrule
      \end{tabular}%
    }
    \label{tab:add-underlying-A}%
  \end{table*}%

Table~\ref{tab:add-underlying-A} shows the online accuracy and test accuracy of different integrated OCL algorithms on the CORe50/ConvNet dataset.
From the table, we can see that both ER, MIR, LWF, and MAS can effectively mitigate catastrophic forgetting, achieving better test accuracy while maintaining comparable online accuracy.
When applying these algorithms to different OCL frameworks, Ferret$_{\mathrm{M+}}$ consistently outperforms the other OCL frameworks by a large margin, demonstrating the effectiveness of Ferret in leveraging memory resources to improve the performance of integrated OCL algorithms.

\begin{figure}
    \centering
    \subfloat[CORe50/ConvNet]{\includegraphics[width=0.5\linewidth]{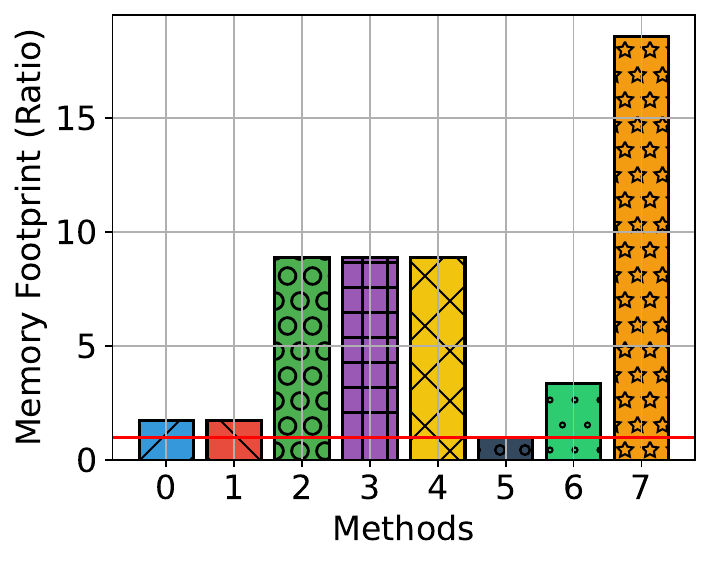}}
    \subfloat[CLEAR100/ResNet]{\includegraphics[width=0.5\linewidth]{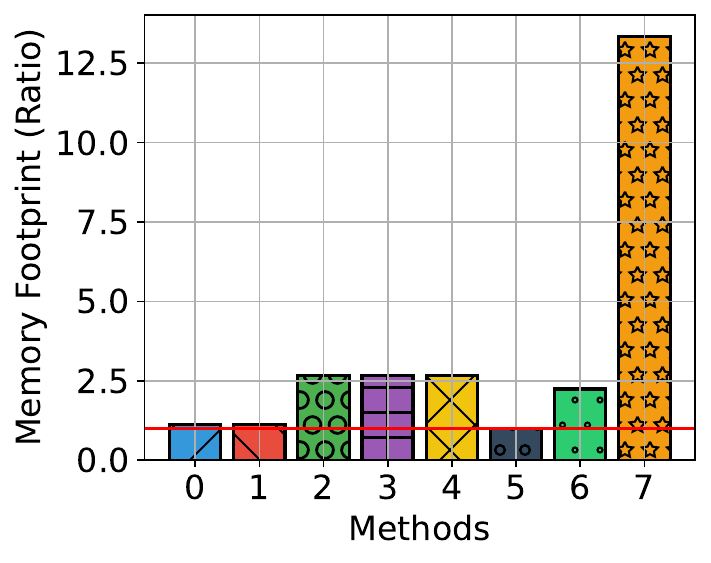}}
    \caption{Consumed memory of different stream learning algorithms. Ferret achieves rapid adaptation across varying memory constraints.}
    \label{fig:add-overall-memory-comparison}
\end{figure}

\begin{figure}
    \centering
    \subfloat[CLEAR100/ResNet]{\includegraphics[width=0.5\linewidth]{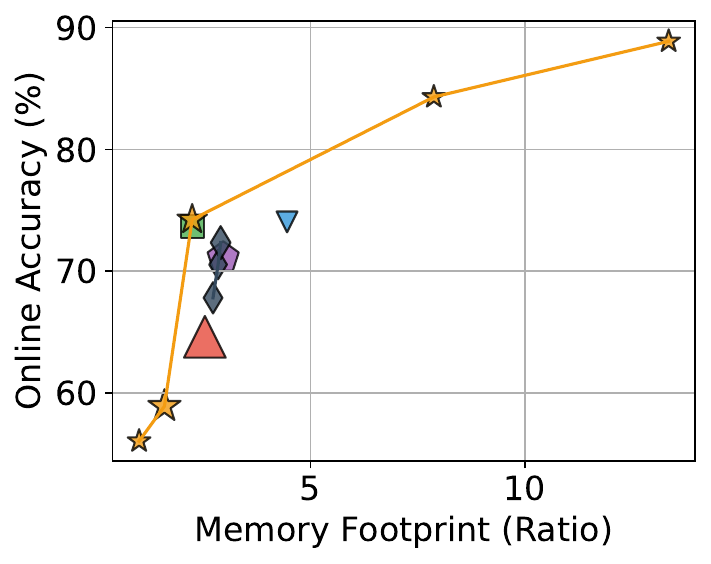}}
    \subfloat[CLEAR100/MobileNet]{\includegraphics[width=0.5\linewidth]{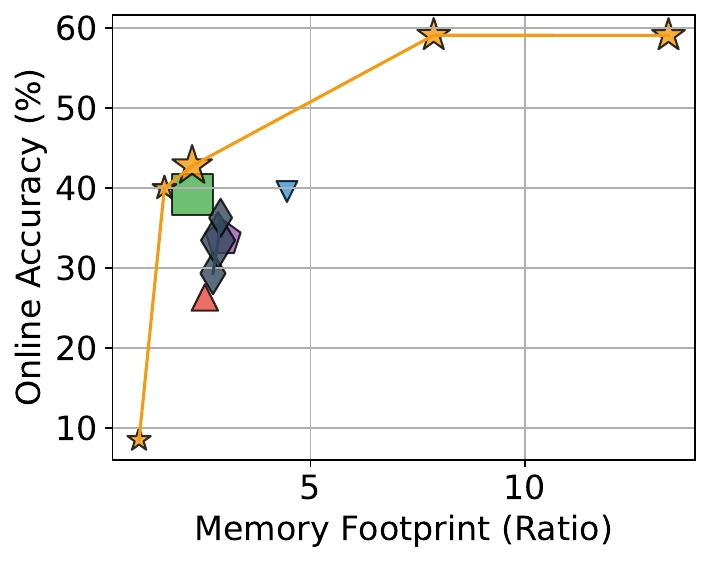}}
    \caption{Relationships between online accuracy and memory consumption of different pipeline parallelism strategies, the marker size represents the standard errors of means.}
    \label{fig:add-pp-memory-comparison}
\end{figure}

Fig.~\ref{fig:add-overall-memory-comparison} and Fig.~\ref{fig:add-pp-memory-comparison} expand results in Fig.~\ref{fig:overall-memory-comparison} and Fig.~\ref{fig:pp-memory-comparison} to show the memory consumption of different stream learning algorithms and the relationship between online accuracy and memory consumption of different pipeline parallelism strategies, respectively.

\section{Limitations}

To reduce the staleness of gradients, the proposed Ferret utilizes Taylor series expansion to approximate the gradient at the current time step.
This compensation introduces an additional hyper-parameter $\lambda$.
However, the optimal $\lambda$ may vary across different datasets and models, which may require manual tuning.
To automate this process, $\lambda$ can be optimized in real-time under the mild assumption that the distributions of $\mathbb{E}_k D^k$ and $\mathbb{E}_k D^{k+1}$ are similar.
This assumption may not hold in scenarios where the data distribution changes significantly over a short time (\textit{e.g.}, emergency handling).
In such cases, $\lambda$ in Ferret requires manual tuning to ensure optimal performance.

\end{document}